\documentclass[runningheads]{llncs}

\usepackage{amssymb}
\usepackage{graphicx}
\usepackage{times}
\usepackage{latexsym}

\usepackage{url}

\usepackage{booktabs} % For formal tables

\usepackage{epstopdf}
\usepackage{multirow}
\usepackage{balance}
\usepackage{amsfonts}
\usepackage{amsmath}
\usepackage{amssymb}
\usepackage{algorithm}
\usepackage{bbold}
\usepackage{algpseudocode}
\usepackage{microtype}
\usepackage{lmodern}
\usepackage{enumitem}

\usepackage{threeparttable}
\usepackage{bm}
\usepackage{subfig}
\usepackage{booktabs}
\usepackage{caption}

\usepackage{url}
\urldef{\mailsa}\path|{alfred.hofmann, ursula.barth, ingrid.haas, frank.holzwarth,|
\urldef{\mailsb}\path|anna.kramer, leonie.kunz, christine.reiss, nicole.sator,|
\urldef{\mailsc}\path|erika.siebert-cole, peter.strasser, lncs}@springer.com|
\newcommand{\keywords}[1]{\par\addvspace\baselineskip
\noindent\keywordname\enspace\ignorespaces#1}

\newcommand{\SetEx}{{SetExpan}}
\newcommand{\SetExNORE}{{SetExpan$^{-re}$}}
\newcommand{\SetExNOCS}{{SetExpan$^{-cs}$}}

\newcommand{\nop}[1]{}

\newtheorem{thm:def}{Definition}
\newtheorem{thm:eg}{Example}
\newtheorem{thm:lem}{Lemma}
\newtheorem{thm:obs}{Observation}

\DeclareMathOperator*{\argmax}{arg\,max}

\begin{document}

\mainmatter  % start of an individual contribution

\title{\SetEx: Corpus-Based Set Expansion via Context Feature Selection and Rank Ensemble}

\titlerunning{\SetEx}

%\author{Alfred Hofmann%
%\thanks{Please note that the LNCS Editorial assumes that all authors have used
%the western naming convention, with given names preceding surnames. This determines
%the structure of the names in the running heads and the author index.}%
%\and Ursula Barth\and Ingrid Haas\and Frank Holzwarth\and\\
%Anna Kramer\and Leonie Kunz\and Christine Rei\ss\and\\
%Nicole Sator\and Erika Siebert-Cole\and Peter Stra\ss er}
% \author{Jiaming Shen$^*$ \and Zeqiu Wu$^*$ \and Dongming Lei \and Jingbo Shang \and Xiang Ren \and Other unknown people \and Jiawei Han}
\author{Jiaming Shen$^\star$, Zeqiu Wu\thanks{Equal Contribution}, Dongming Lei, Jingbo Shang, Xiang Ren, Jiawei Han}
\authorrunning{J. Shen, Z. Wu, D. Lei, J. Shang, X. Ren, J. Han}

% \institute{Computer Science Department, University of Illinois at Urbana-Champaign, Urbana, IL 61801, USA\\
\institute{Department of Computer Science, University of Illinois at Urbana-Champaign, USA\\[.5ex]
\{js2, zeqiuwu1, dlei5, shang7, xren7, hanj\}@illinois.edu\\
}


\toctitle{Lecture Notes in Computer Science}
\tocauthor{Authors' Instructions}
\maketitle

\begin{abstract}
	%!TEX root = main.tex
% UTF-8 encoding
\emph{Corpus-based set expansion} (i.e., finding the ``complete" set of entities belonging to the same semantic class, based on a given corpus and a tiny set of seeds) is a critical task in knowledge discovery.
It may facilitate numerous downstream applications, such as information extraction, taxonomy induction, question answering, and web search.

To discover new entities in an expanded set, previous approaches either make \emph{one-time entity ranking} based on distributional similarity, or resort to \emph{iterative pattern-based bootstrapping}.
The core challenge for these methods is how to deal with noisy context features derived from free-text corpora, which may lead to entity intrusion and semantic drifting.
In this study, we propose a novel framework, \emph{\SetEx}, which tackles this problem, with two techniques:
(1) a context feature selection method that selects clean context features for calculating entity-entity distributional similarity, and
(2) a ranking-based unsupervised ensemble method for expanding entity set based on denoised context features.
Experiments on three datasets show that \emph{\SetEx} is robust and outperforms previous state-of-the-art methods in terms of mean average precision.

\keywords{Set Expansion, Information Extraction, Bootstrapping, Unsupervised Ranking-Based Ensemble} 
\end{abstract}

%!TEX root = main.tex
% UTF-8 encoding
\section{Introduction}

Set expansion refers to the problem of expanding a small set of seed entities into a complete set of entities that belong to the same semantic class \cite{Wang2007LanguageIndependentSE}.
For example, if a given seed set is
\{\emph{Oregon, Texas, Iowa}\},
set expansion should return a hopefully complete set of entities in the same semantic class, ``\emph{U.S. states}".
Set expansion can benefit various downstream applications, such as knowledge extraction \cite{gupta2014improved}, taxonomy induction \cite{velardi2013ontolearn}, and web search \cite{chen2016long}.

One line of work for solving this task includes \emph{Google Set} \cite{tong2008system}, \emph{SEAL} \cite{Wang2007LanguageIndependentSE}, and \emph{Lyretail} \cite{chen2016long}.
In this approach, a query consisting of seed entities is submitted to a search engine to mine top-ranked webpages.
While this approach can achieve relatively good quality, the required seed-oriented online data extraction is costly. 
Therefore, more studies \cite{Pantel2009WebScaleDS}\cite{Shi2010CorpusbasedSC}\cite{He2011SEISASE}\cite{Wang2015ConceptEU}\cite{rong2016egoset} are proposed in a \emph{corpus-based} setting where sets are expanded by offline processing based on a specific corpus.

For \emph{corpus-based} set expansion, there are two general approaches, \emph{one-time entity ranking} and \emph{iterative pattern-based bootstrapping}.
Based on the assumption that similar entities appear in similar contexts, the first approach \cite{Pantel2009WebScaleDS}\cite{Shi2010CorpusbasedSC}\cite{He2011SEISASE} makes a one-time ranking of candidate entities based on their distributional similarity with seed entities.
A variety of ``contexts'' are used, including Web table, Wikipedia list, or just free-text patterns, and entity-entity distributional similarity is calculated based on \emph{all} context features.
However, blindly using \emph{all} such features can introduce undesired entities into the expanded set because many context features are not representative for defining the target semantic class although they do have connections with some of the seed entities.
For example, when expanding the seed set \{\emph{Oregon, Texas, Iowa}\}, ``\emph{located in $\underline{\hspace{0.1in}}$}" can be a pattern feature (the entity is replaced with a placeholder) strongly connected to all the three seeds. 
However, it does not clearly convey the semantic meaning of \emph{``U.S. states."} and can bring in entities like \emph{USA} or \emph{Ontario} when being used to calculate candidate entity's similarity with seeds. 
This is \emph{entity intrusion} error.
Another issue with this approach is that it is hard to obtain the full set at once without back and forth refinement. 
In some sense, iteratively bootstrapped set expansion is a more conservative way and leads to better precision.

The second approach, iterative pattern-based bootstrapping \cite{Shi2014APC}\cite{gupta2014improved}\cite{Gupta2015DistributedRO}, starts from seed entities to extract quality patterns, based on a predefined pattern scoring mechanism, and it then applies extracted patterns to obtain even higher quality entities using another entity scoring method.
This process iterates and the high-quality patterns from all previous iterations are accumulated into a pattern pool which will be used for the next round of entity extraction.
This approach works only when patterns/entities extracted at each iteration are highly accurate, otherwise, it may cause severe \emph{semantic shift} problem.
Suppose in the previous example, \emph{``located in $\underline{\hspace{0.1in}}$"} is taken as a good pattern from the seed set \emph{\{Oregon, Texas, Iowa\}}, and
this pattern brings in \emph{USA} and \emph{Ontario}.
These undesired entities may bring in even lower quality patterns and iteratively cause the set shifting farther away.
Thus, the pattern and entity scoring methods are crucial but sensitive in iterative bootstrapping methods. 
If they are not defined perfectly, the semantic shift can cause big problems. 
However, it is hard to have a perfect scoring mechanism due to the diversity and noisiness of unstructured text data.

\begin{figure}[!thbp]
	\centerline{\includegraphics[width=1.00\textwidth]{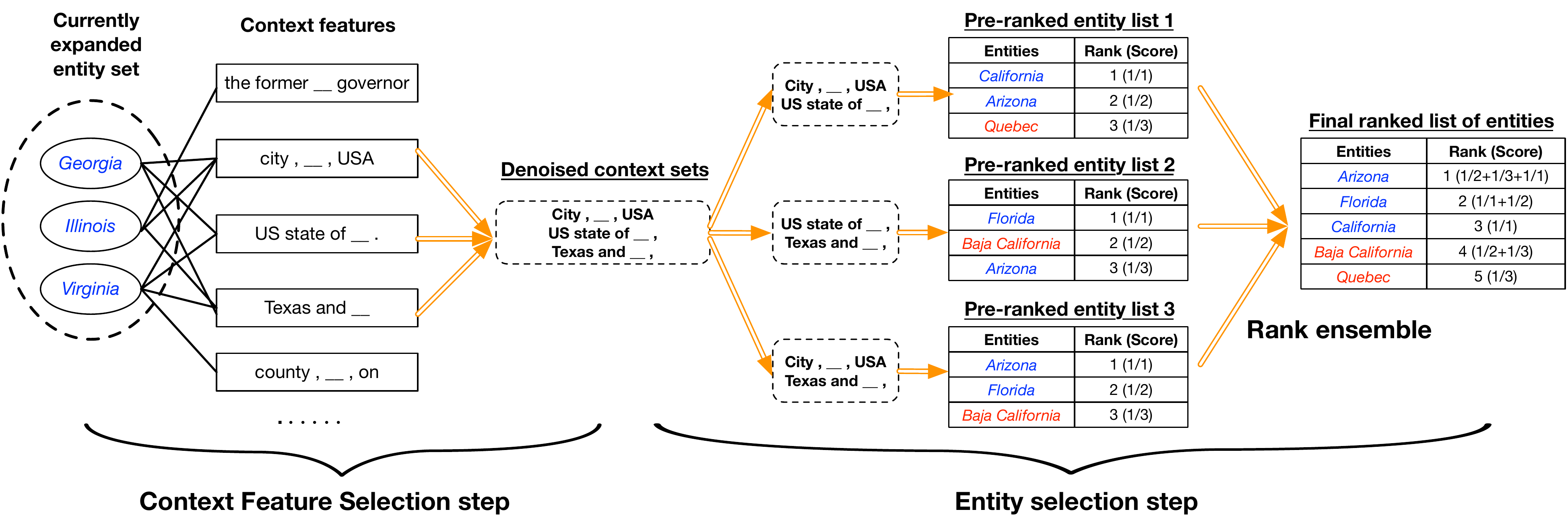}}
	\caption{An example showing two steps in one iteration of \SetEx.}
	\label{fig:overallFramework}
\end{figure}

This study proposes a new set expansion framework, \SetEx, which addresses both challenges posed above for corpus-based set expansion on free text.
It carefully and conservatively extracts each candidate entity and iteratively improves the results.
First, to overcome the entity intrusion problem, instead of using all context features, context features are carefully selected by calculating distributional similarity.
Second, to overcome the semantic drift problem, different from other bootstrapped approaches, our high-quality feature pool will be reset at the beginning of each iteration.
Finally, our carefully designed unsupervised ranking-based ensemble method is used at each iteration to further refine entities and make our system robust to noisy or wrongly extracted pattern features.

Figure \ref{fig:overallFramework} shows the pipeline at each iteration. 
\SetEx~iteratively expands an entity set through a context feature selection step and an entity selection step. 
At the context feature selection, each context feature is  scored based on its strength with currently expanded entities and top-ranked context features are selected. 
At the entity selection step, multiple subsets of the selected representative context features are sampled and each subset is used to obtain a ranked entity list. 
Finally, all the ranked lists are collected to compute the final ranking list of each candidate entity for expansion.

The major contributions of this paper are: 
(1) we propose an iterative set expansion framework with a novel context feature selection approach, to handle the issues of entity intrusion and semantic drift; 
(2) we develop an unsupervised ranking-based ensemble algorithm for entity selection to make our system robust and further reduce the impact of semantic drift.
To evaluate the \SetEx~method, we use three publicly available datasets and manually label expanded results of 65 queries over 13 semantic classes. 
Empirical results show that \SetEx~outperforms the state-of-the-art baselines in terms of Mean Average Precision. 
Code\footnote{\scriptsize \url{https://github.com/mickeystroller/SetExpan}} and datasets\footnote{\scriptsize \url{https://tinyurl.com/SetExpan-data}} described in this paper are publicly. 
%!TEX root = main.tex
% UTF-8 encoding
\vspace{-1ex}
\section{Related Work}
\vspace{-1ex}
The problem of completing an entity set given several seed entities has attracted extensive research efforts due to its practical importance. 
Google Sets \cite{tong2008system} was among the earliest work dealing with this problem.
It used proprietary algorithms and is no longer publicly accessible. 
Later, Wang and Cohen proposed \emph{SEAL} system \cite{Wang2007LanguageIndependentSE}, which first submits a query consisting of all seed entities into a general search engine and then mines the top-ranked webpages. 
Recently, Chen et al.\ \cite{chen2016long} improved this approach by leveraging a ``page-specific'' extractor built in a supervised manner and showed good performance on long-tail (i.e., rare) term expansion.
All these methods need an external search engine and require seed-oriented data extraction. 
In comparison, our approach conducts corpus-based set expansion without resorting to online data extraction from specific webpages.

To tackle the corpus-based set expansion problem, Ghahramani and Heller \cite{Ghahramani2005BayesianS} used a Bayesian method to model the probability that a candidate entity belongs to some unknown cluster that contains the input seeds. 
Pantel et al.\ \cite{Pantel2009WebScaleDS} developed a web-scale set expansion pipeline by exploiting distributional similarity on context words for each candidate entity. 
He et al.\ proposed the SEISA system \cite{He2011SEISASE} that used query logs along with web lists as external evidence besides free text, and designed an iterative similarity aggregation function for set expansion.
Recently, Wang et al. \cite{Wang2015ConceptEU} leveraged web tables and showed very competitive results when not only seed entities but also intended class name were given. 
While these semi-structured lists and tables are helpful, they are not always available for some specific domain corpus such as PubMed articles or DBLP papers. 
Perhaps the most relevant work to ours is by Rong \cite{rong2016egoset}. 
In that paper, the authors used the skip-gram feature combined with additional user-generated ontologies (i.e., Wikipedia list) for set expansion. 
However, they targeted the multifaceted expansion and exploited all skip-gram features for calculating the similarity because two entities. 
In our work, we keep the core idea of distributional similarity but calculate such similarity using only carefully selected \emph{denoised} context features.

In a broader sense, our work is also related to information extraction and named entity recognition. Without given enough training data, bootstrapped entity extraction system \cite{Etzioni2005UnsupervisedNE}\cite{Gupta2014ResearchAA}\cite{gupta2014improved} is the most popular and effective choice. 
At each bootstrap iteration, the system will first create patterns around entities; score patterns based on their ability to extract more positive entities and less negative entities (if provided), and use top-ranked patterns to extract more candidate entities. Multiple pattern scoring and entity scoring functions are proposed.
For example, Riloff et al.\ \cite{Riloff1996AutomaticallyGE} scored each pattern by calculating the ratio of positive entities among all entities extracted by it, and scored each candidate entity by the number and quality of its matched patterns. 
Gupta et al.\ \cite{Gupta2014ResearchAA} scored patterns using the ratio of scaled frequencies of positive entities among all entities extracted by it. All these methods are heuristic and sensitive to different model parameters.

More generally, our work is also related to class label acquisition \cite{Talukdar2008WeaklySupervisedAO}\cite{Wang2009SemisupervisedLO} which aims to propagate class labels to data instances based on labeled training examples, and entity clustering \cite{Balasubramanyan2013FromTM}\cite{Lin2009PhraseCF} where the goal is to find clusters of entities. 
However, the class label acquisition methods require a much larger number of training examples than the typical size of user input seed set, and the entity clustering algorithms can only find semantically related entities instead of entities strictly in the same semantic class.

\vspace{-1ex}
%!TEX root = main.tex
% UTF-8 encoding
%% Our Methodology

\section{Our Methodology: The \SetEx~Framework}
This section introduces first the context features and data model used by \SetEx\ in Sect.\ \ref{subsec:cfdm} and then our context-dependent similarity measure in Sect.\ \ref{subsec:cds}.
It then discusses how to select context features in Sect.\ \ref{subsec:cd} and presents our novel unsupervised ranking-based ensemble method for entity selection in Sect.\ \ref{subsec:re}.

\vspace{-1.0ex}
\smallskip
\subsection{Data Model and Context Features}\label{subsec:cfdm}
\begin{figure}[!t]
	\centerline{\includegraphics[width=0.60\textwidth]{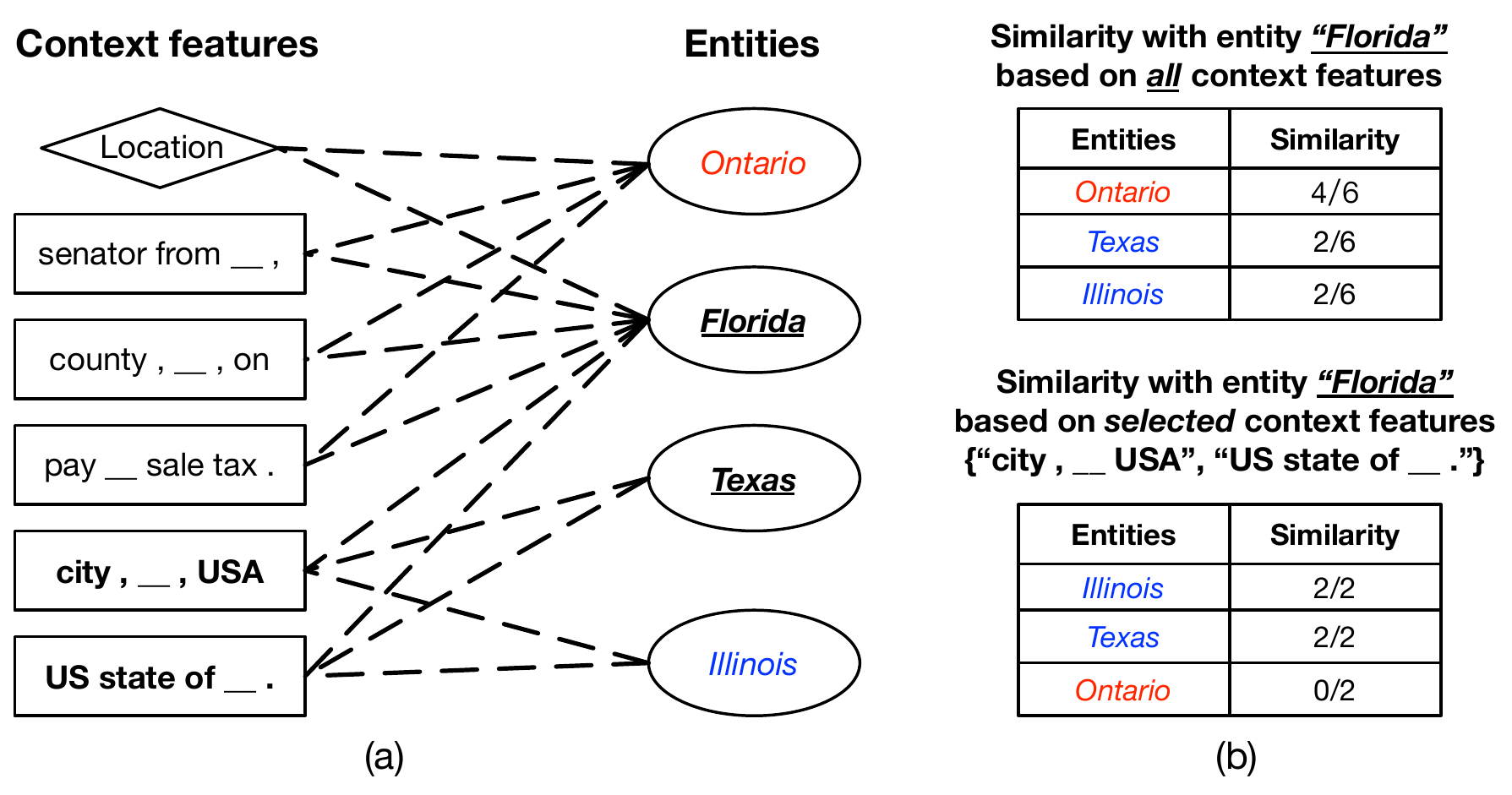}}
	\caption{(a) A simplified bipartite graph data model. (b) Similarity with seed entity conditioned on two different sets of context features.}
	\label{fig:datamodel}
\end{figure}

We explore two types of context features obtained from the plain text: (1) skip-grams \cite{rong2016egoset} and (2) coarse-grained types \cite{gupta2014improved}.
Data is modeled as a bipartite graph (Figure \ref{fig:datamodel}(a)), with candidate entities on one side and their context features on the other.
Each type of context features are described as follows.

\smallskip \noindent
\textbf{Skip-gram:} Given a target entity $e_{i}$ in a sentence, one of its skip-gram is ``$w_{-1}$ $\underline{\hspace{0.1in}}$ $w_{1}$'' where $w_{-1}$ and $w_{1}$ are two context words and $e_{i}$ is replaced with a placeholder.
For example, one skip-gram of entity \emph{``Illinois"} in sentence \emph{``We need to pay Illinois sales tax."} is ``pay $\underline{\hspace{0.1in}}$ sales".
As suggested in \cite{rong2016egoset}, we extract up to six skip-grams of different lengths for one target entity $e_{i}$ in each sentence.
One advantage of using skip-grams is that it imposes strong positional constraints.

\smallskip
\noindent
\textbf{Coarse-grained type:} Besides the unstructured skip-gram features, we use coarse-grained type to filter those obviously-wrong entities.
For examples, when we expand the ``U.S. states'', we will not consider any entity that is typed ``Person''.
After this process, we can obtain a cleaner subset of candidate entities.
This mechanism is also adopted in \cite{gupta2014improved}.

\smallskip

After obtaining the ``nodes" in bipartite graph data model, we need to model the edges in the graph.
In this paper, we assign the weight between each pair of entity $e$ and context feature $c$ using the \textit{TF-IDF transformation} \cite{rong2016egoset}, which is calculated as follows:
\begin{equation}
\small
f_{e,c} = \log(1+X_{e,c}) \left[\log|E| - \log\left(\sum_{e'} X_{e',c}\right) \right],
\end{equation}
where $X_{e,c}$ is the raw co-occurrence count between entity $e$ and context feature $c$, $|E|$ is the total number of candidate entities.
We refer to such scaling as the \textit{TF-IDF transformation} since it resembles the \textit{tf-idf} scoring in information retrieval if we treat each entity $e$ as a ``document'' and each of its context feature $c$ as a ``term''.
Empirically, we find such weight scaling performs outperforms some other alternatives such as point-wise mutual information (PMI) \cite{He2011SEISASE}, truncated PMI \cite{McIntosh2008WeightedME}, and BM25 scoring \cite{Ren2017ComparativeDA}.
\vspace{-0.5ex}
\subsection{Context-dependent Similarity}\label{subsec:cds}
With the bipartite graph data model constructed, the task of expanding an entity set at each iteration can be viewed as finding a set of entities that are most ``similar'' to the currently expanded set.
In this study, we use the weighted Jaccard similarity measure.
Specifically, given a set of context features $F$, we calculate the \emph{context-dependent} similarity as follows:
\begin{equation}
\small
Sim(e_1, e_2 | F) = \frac{\sum_{c \in F} \min(f_{e_1,c}, f_{e_2, c}) }{\sum_{c \in F} \max(f_{e_1,c}, f_{e_2, c}) }.
\end{equation}
Notice that if we change context feature set $F$, the similarity between entity pair is likely to change, as demonstrated in the following example.
\smallskip
\begin{thm:eg}\label{example3} % context-dependent similarity
Figure \ref{fig:datamodel}(a) shows a simplified bipartite graph data model where all edge weights are equal to 1 (and thus omitted from the graph for clarity).
The entity ``\textsf{Florida}" connects with all 6 different context features, while the entity ``\textsf{Ontario}" is associated with top 4 context features including 1 type feature and 3 skip-gram features.
If we add all the 6 possible context features into the context feature set $F$, the similarity between ``\textsf{Florida}" and ``\textsf{Ontario}" is $\frac{1+1+1+1}{1+1+1+1+1+1} = \frac{4}{6}$.
On the other hand, if we put only two context features ``city , $\underline{\hspace{0.1in}}$, USA'', ``US state of $\underline{\hspace{0.1in}}$ .'' into $F$, the similarity between same pair of entities will change to $\frac{1+1}{1+1} = \frac{2}{2}$.
Therefore, we refer such similarity as context-dependent similarity.
\end{thm:eg}
Finally, we want to emphasize that our proposed method is general in the sense that other common similarity metrics such as cosine similarity can also be used.
In practice, we find the performance of a set expansion method depends not really on the exact choice of base similarity metrics, but more on which contexts are selected for calculating \emph{context-dependent} similarity.
Similar results were also reported in a previous study \cite{He2011SEISASE}.

\vspace{-0.5ex}
\subsection{Context Feature Selection}\label{subsec:cd}

As shown in Example \ref{example3}, the similarity between two entities really depends on the selected feature set $F$.
The motivation of context feature selection is to find a feature subset $F^{*}$ of fixed size $Q$ that best ``profiles" the target semantic class.
In other words, we want to select a feature set $F^{*}$ based on which entities within target class are most ``similar" to each other.
Given such $F^{*}$, the entity-entity similarity conditioned on it can best reflect their distributional similarity with regard to the target class.
In some sense, such $F^{*}$ best profiles the target semantic class.
Unfortunately, to find such $F^{*}$ of fixed size $Q$, we need to solve the following optimization problem which turns out to be NP-Hard, as shown in \cite{Chierichetti2010FindingTJ}.

\begin{equation}
\small
F^{*} = \argmax_{|F| = Q} \sum_{i=1}^{|X|}\sum_{j>i}^{|X|} Sim(e_{i}, e_{j} | F),
\end{equation}
where $X$ is the set of currently expanded entities. Initially, we treat the user input seed set $S$ as $X$. As iterations proceed, more entities will be added into $X$.

Given the NP-hardness of finding the optimal context feature set, we resort to a heuristic method that first scores each context feature based on its accumulated strength with entities in $X$ and then selects top $Q$ features with maximum scores.
This process is illustrated in the following example:
\smallskip
\begin{thm:eg}\label{example4} % context denoising
For demonstration purpose, we again assume all edge weights in Figure \ref{fig:datamodel}(a) are equal to 1 and let the currently expanded entity set $X$ be \{``\textsf{Florida}'', ``\textsf{Texas}''\}.
Suppose we want to select two ``denoised'' context features, we will first score each context feature based on its associated entities in $X$.
The top 4 contexts will obtain a score 1 since they match only one entity in $X$ with strength 1, and the 2 contexts below will get a score 2 because they match both entities in $X$.
Then, we rank context features based on their scores and select 2 contexts with highest scores: ``city , $\underline{\hspace{0.1in}}$, USA'', ``US state of $\underline{\hspace{0.1in}}$ .'' into $F$.
\end{thm:eg}

Finally, we want to emphasize two major differences of our context feature selection method from other heuristic ``pattern selection" methods.
First, most pattern selection methods require either users to explicitly provide the ``negative'' examples for the target semantic class \cite{Jindal2011LearningFN}\cite{gupta2014improved}\cite{Shi2014APC}, or implicitly expand multiple mutually exclusive classes in which instances in one class serve as negative examples for all the other classes \cite{Curran2007MinimisingSD}\cite{McIntosh2008WeightedME}.
Our method requires only a small number of ``positive'' examples.
In most cases, it is hard for humans to find good discriminative negative examples for one class, or to provide both mutually exclusive and somehow related comparative classes.
Second, the bootstrapping method will add its selected ``quality patterns'' during each iteration into a quality pattern pool, while our method will select high quality context features at each iteration from scratch.
If one noisy pattern is selected and added into the pool, it will continue to introduce more irrelevant entities at all the following iterations.
Our method can avoid such noise accumulation.

\vspace{-0.5ex}
\subsection{Entity Selection via Rank Ensemble}\label{subsec:re}

\begin{figure}[!t]
	\centerline{\includegraphics[width=0.90\textwidth]{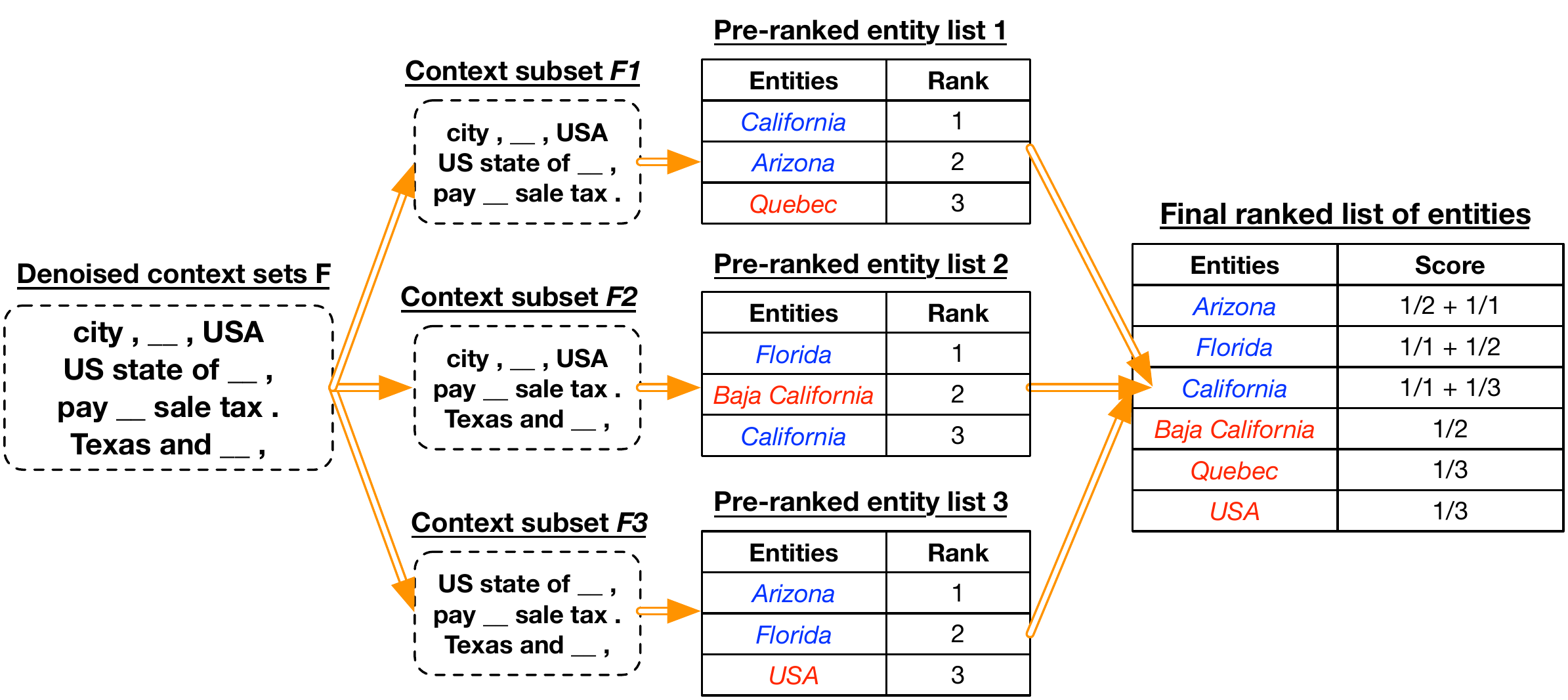}}
	\caption{A toy example to show entity selection via rank ensemble.}
	\label{fig:re}
\end{figure}

Intuitively, the entity selection problem can be viewed as finding those entities that are most similar to the currently expanded set $X$ conditioned on the selected context feature set $F$.
To achieve this, we can rank each candidate entity based on its score in eq. (\ref{eq:entityscore}) and then add top-ranked ones into the expanded set:
\begin{equation}\label{eq:entityscore}
\small
score(e|X,F) = \frac{1}{|X|} \sum_{e' \in X} Sim(e, e' | F).
\end{equation}
However, due to the ambiguity of natural language in free-text corpora, the selected context feature set $F$ may still be noisy in the sense that an irrelevant entity is ranked higher than a relevant one. To further reduce such errors, we propose a novel ranking-based ensemble method for entity selection.

The key insight of our method is that an inferior entity will not appear frequently in multiple pre-ranked entity lists at top positions.
Given a selected context set $F$, we first use sampling without replacement method to generate $T$ subsets of context features $F_{t}, t = 1, 2, \ldots, T$.
Each subset is of size $\alpha|F|$ where $\alpha$ is a model parameter within range $[0,1]$.
For each $F_{t}$, we can obtain a pre-ranked list of candidate entities $L_{t}$ based on $score(e | X, F_{t})$ defined in eq. (\ref{eq:entityscore}).
We use $r_{t}^{i}$ to denote the rank of entity $e_{i}$ in list $L_{t}$.
If entity $e_{i}$ does not appear in $L_{t}$, we let $r_{t}^{i} = \infty$.
Finally, we collect $T$ pre-ranked lists and score each entity based on its mean reciprocal rank (\emph{mrr}).
All entities with average rank above $r$, namely $mrr(e) \leq T/r$, will be added into entity set $X$.
\begin{equation}\label{eq:mrrscore}
\small
mrr(e_{i}) = \sum_{t=1}^{T} \frac{1}{r_{t}^{i}},  \qquad r_{t}^{i} = \sum_{e_{j} \in E} I\left(score(e_{i} | X, F_{t}) \leq score(e_{j} | X, F_{t}) \right),
\end{equation}
where $I(\cdot)$ is the indicator function.
Naturally, a relevant entity will rank at top position in multiple pre-ranked lists and thus accumulate a high \emph{mrr} score, while an irrelevant entity will not consistently appear in multiple lists at high position which leads to low \emph{mrr} score.
Finally, we use the following example to demonstrate the whole process of entity selection.
\smallskip
\begin{thm:eg}\label{example5} % rank ensemble
In Figure \ref{fig:re}, we want to expand the ``US states'' semantic class given a selected context feature set $F$ with 4 features.
We first sample a subset of 3 context features $F_{1}$ = \{``city , $\underline{\hspace{0.1in}}$ , USA", ``US state of $\underline{\hspace{0.1in}}$ ,'', ``pay $\underline{\hspace{0.1in}}$ sales tax ."\}, and then use $F_{1}$ to obtain a pre-ranked entity list $L_{1} =$ $\langle``\textsf{California}", ``\textsf{Arizona}", ``\textsf{Quebec}"\rangle$.
By repeating this process three times, we get 3 pre-ranked lists and ensemble them into a final ranked list in which entity ``\textsf{Arizona}" is scored 1.5 because it is ranked in the 2nd position in $L_{1}$ and 1st position in $L_{3}$.
Finally, we add those entities with mrr score larger than 1, meaning this entity is ranked at 3rd position on average, into the expanded set $X$.
In this simple example, the model parameters $T=3, \alpha = \frac{|F_{1}|}{|F|} = 0.75$, and $r = 3$.
\end{thm:eg}

    \begin{algorithm}[!t]
    \scriptsize
    \caption{\SetEx}\label{alg:SetEx}
    \begin{algorithmic}[1]
    	%% Initialization
    	\State \textbf{Input:} Candidate entity set $E$, initial seed set $S$, entity-context graph $G$, expected size of output set $K$, model parameters $\{Q, T, \alpha, r \}$.
	\State \textbf{Output:} The expanded set $X$.
	\State $X$ = $S$.
	%% Update
	\While{$|X| \leq K$}
		\State Set $F = \emptyset$ // Select denoised contexts from scratch
		\State Score context features based on $X$ and add top $Q$ denoised contexts into $F$.
		\State // Entity-selection via rank ensemble
		\For{$t = 1,2,\ldots, T$}
			\State Uniformly sample $\alpha Q$ contexts and construct feature subset $F_{t}$.
			\State Score entities based on Eq. (\ref{eq:entityscore}) given $F_{t}$ and obtain the pre-ranked list $L_{t}$.
			\State Update the $mrr$ score of each entity based on Eq. (\ref{eq:mrrscore}).
		\EndFor
		\State $X = X \cup \{e| mrr(e) \ge \frac{T}{r}\}$ // Add entities into expanded set $X$ .
	\EndWhile
	\State Return $X$.
        \end{algorithmic}
    \end{algorithm}
    
% \subsection{Summary}
\smallskip \noindent
\textbf{Put all together.}
Algorithm \ref{alg:SetEx} summarizes the whole \SetEx~process. 
The candidate entity set $E$ and bipartite graph data model $G$ are pre-calculated and stored. 
A user needs only to specify the seed set $S$ and the expected size of output set $K$.
There is a total of 4 model parameters: the number of top quality context features selected in each iteration $Q$, the number of pre-ranked entity lists $T$, the relative size of feature subset $0 < \alpha < 1$, and final $mrr$ threshold $r$.
The tuning and sensitivity of these parameters will be discussed in the experiment section.

%!TEX root = main.tex
% UTF-8 encoding
\vspace{-0.5ex}
\section{Experiments}

%Experimental Setup
\subsection{Experimental Setup}
%% Datasets preparation
\vspace{-1.0ex}
\subsubsection{Datasets preparation.}

\emph{\SetEx} is a \emph{corpus-based} entity set expansion system and thus we use three corpora to evaluate its performance.
Table \ref{tab:dataset} lists 3 datasets we used in experiments. (1) \textbf{APR} is constructed by crawling all 2015 news articles from AP and Reuters. (2) \textbf{Wiki} is a subset of English Wikipedia used in \cite{Ling2012FineGrainedER}. (3) \textbf{PubMed-CVD} is a collection of research paper abstracts about cardiovascular disease retrieved from PubMed.

For APR and PubMed-CVD datasets, we adopt a data-driven phrase mining tool \cite{liu2015mining} to obtain entity mentions and type them using ClusType \cite{ren2015clustype}.
Each entity mention is mapped heuristically to an entity based on its lemmatized surface name.
We then extract variable-length skip-grams for all entity mentions as features for their corresponding entities, and construct the bipartite graph data model as introduced in the previous section.
For Wiki dataset, the entities have already been extracted and typed using distant supervision.
For the type information in each dataset, there are 16 coarse-grained types in APR and 4 coarse-grained types in PubMed-CVD.
For Wiki, since it originally has about 50 fine-grained types, which may reveal too much information, we manually mapped them to 11 more coarse-grained types.
%% Query construction

\textbf{Query construction.}
A query is a set of seed entities of the same semantic class in a dataset, serving as the input for each system to expand the set. 
The process of query generation is as follows. For each dataset, we first extract 2000 most frequent entities in it and construct an entity list. 
Then, we ask three volunteers to manually scan the entity lists and propose a few semantic classes for each list. The proposed class should be interesting, relatively unambiguous and has a reasonable coverage in its corresponding corpus. 
These semantic classes cover a wide variety of topics, including locations, companies as well as political parties, and have different degrees of difficulty for set expansion. 
After finalizing the semantic classes for each dataset, the students randomly select entities of each semantic class from the frequent entity list to form 5 queries of size 3.
To select the queries for PubMed-CVD, we seek help from two additional students with biomedical expertise, following the same previous approach.
Due to the large size of PubMed-CVD dataset and runtime limitation, we only select 1 semantic class (hormones) with 5 queries.

With all queries selected, we have humans to label all the classes and instances returned by each of the following 7 compared methods. For APR and Wiki datasets, the inter-rater agreements (kappa-value) over three students are 0.7608 and 0.7746, respectively. For PubMed-CVD dataset, the kappa-value is 0.9236. All entities with conflicting label results are further resolved after discussions among all human labelers. Thus, we have our ground truth datasets.

\begin{table}[!t]
\scriptsize
\centering
\caption{Datasets statistics and Query descriptions}
\begin{tabular}{|c|c|c|c|c|c|}
\hline
Dataset & FileSize & \#Sentences & \#Entities & \#Test queries\\
\hline\hline
APR & 775MB & 1.01M & 122K & 40 \\
Wiki & 1.02GB & 1.50M & 710K & 20 \\
PubMed-CVD & 9.3GB & 23M & 179K & 5 \\
\hline
\end{tabular}
\label{tab:dataset}
\end{table}

\textbf{Compared methods.}
Since the focus on this work is the corpus-based set expansion, we do not compare with other methods that require online data extractions. Also, to further analyze the effectiveness of each module in \SetEx~framework. We implement 3 variations of our framework.
\vspace{-1.5ex}
\begin{itemize}
\item word2vec \cite{Mikolov2013DistributedRO}: We use the ``skip-gram'' model in word2vec to learn the embedding vector for each entity, and then return $k$ nearest neighbors around seed entities as the expanded set.
\item PTE \cite{tang2015line5pte}: We first construct a heterogeneous information network including entity, skip-gram features, and type features. PTE model is then applied to learn the entity embedding which is used to determine the $k$ nearest neighbors around seed entities.
\item SEISA \cite{He2011SEISASE}:
An entity set expansion algorithm based on iterative similarity aggregation. It uses the occurrence of entities in web list and query log as entity features. In our experiments, we replace the web list and query log with our skip-gram and coarse-grained context features.
\item EgoSet \cite{rong2016egoset}:
A multifaceted set expansion system based on skip-gram features, word2vec embeddings and WikiList.
The original system is proposed to expand a seed set to multiple entity sets, considering the ambiguities in seed set.
To achieve this, we use a community detection method to separate the extracted entities into several communities.
However, in order to better compare with EgoSet, we carefully select queries that have little ambiguity or at least the seed set in the query is dominating in one semantic class.
Thus, we discard the community detection part in EgoSet and treat all extracted entities as in one semantic class.
\item \SetExNOCS: Disable the context feature selection module in \SetEx, and use all context features to calculate distributional similarity.
\item \SetExNORE: Disable the rank ensemble module in \SetEx. Instead, we use all selected context feature to rank candidate entities at one time and add top-ranked ones into the expanded set.
\item \SetEx$^{\textit{full}}$: The full version of our proposed method, with both context feature selection and rank ensemble components enabled.
\end{itemize}
\vspace{-1.0ex}
For fair comparison, we try different combinations of parameters and report the best performance for each baseline method. 

%% Evaluation Metrics
\textbf{Evaluation Metrics.}
For each test case, the input is a query, which is a set of 3 seed entities of the same semantic class. The output will be a ranked list of entities. 
For each query, we use the conventional average precision AP$_{k}(c, r)$ at $k$ ($k$ = 10, 20, 50) for evaluation, given a ranked list of entities $c$ and an unordered ground-truth set $r$. 
For all queries under a semantic class, we calculate the mean average precision (MAP) at $k$ as $\frac{1}{N}\sum_{i} AP_{k}(c_{i},r)$, where N is the number of queries. 
To evaluate the performance of each approach on a specific dataset, we calculate the mean-MAP (MMAP) at $k$ over all queried semantic classes as MMAP$_k$ = $\frac{1}{T}\sum_{t=1}^{T} [(\frac{1}{N_{t}}) \sum_{i} AP_{k}(c_{ti},r_{t}) ]$, where $T$ is the number of semantic classes, $N_{t}$ is the number of queries of $t$-th semantic class, $c_{ti}$ is the extracted entity list for $i$-th query for $t$-th semantic class, and $r_{t}$ is the ground truth set for $t$-th semantic class.

\subsection{Experimental Results}

 %% Baselines
\begin{table}[!t]
\centering
\scriptsize
\caption{Overall end-to-end performance evaluation on 3 datasets over all queries.}
\begin{tabular}{|c|c|c|c|c|c|c|c|c|c|}
\hline
\multirow{2}{*}{Methods}  & \multicolumn{3}{c|}{APR} & \multicolumn{3}{c|}{Wiki} & \multicolumn{3}{c|}{PubMed-CVD} \\ \cline{2-10}
& \tiny MAP@10  & \tiny MAP@20 & \tiny MAP@50 & \tiny MAP@10  & \tiny MAP@20  & \tiny MAP@50 & \tiny MAP@10      & \tiny MAP@20      &\tiny  MAP@50      \\
\hline
\hline
EgoSet   &   0.3949            &   0.3942     &   0.3706           &   0.5899     &  0.5754       &    0.5622    &      0.0511      &  0.0410     &  0.0441           \\
\hline
SEISA      &   0.7423          &    0.6090    &   0.3892          &    0.7643    &   0.6606      &   0.4998     &      -      &    -   &     -        \\
\hline
word2vec         &    0.6054    &    0.5385    &    0.4180    &  0.7193      &   0.6289      &   0.4510     &     0.8427       &   0.7701    &   0.6895          \\
\hline
PTE               &    0.3144    &    0.2777    &     0.1996   &     0.6817   &       0.5596  &  0.3839      &      0.9071      &     0.7654  &       0.5641      \\
\hline
\hline
\SetEx$^{-cs}$              &   0.8240     &   0.7997     &   0.7674     &    0.9540    &   0.8955      &   0.7439     &    \textbf{1.000}        &  \textbf{1.000}    &   0.5991          \\
\hline
\SetEx$^{-re}$               &  0.8509      &   0.7792     &  0.7681       &    0.9392    &    0.8680     &   0.7291     &   \textbf{1.000}        &   0.9605    &    0.7371         \\
\hline
\SetEx$^{\textit{full}}$         &     \textbf{0.8967}   &    \textbf{0.8621}    &   \textbf{0.7885}     &    \textbf{0.9571}    &   \textbf{0.9010}      &   \textbf{0.7457}     &   \textbf{1.000}         &  \textbf{1.000}     &    \textbf{0.7454}         \\
\hline
\end{tabular}
\label{tab:datasetMMAP}
\end{table}

%% Overall results on each concept class
\begin{figure}[t]
\centering
    \subfloat[\small APR Country \label{subfig:three-types-links}]{%
      \includegraphics[width=0.33\textwidth]{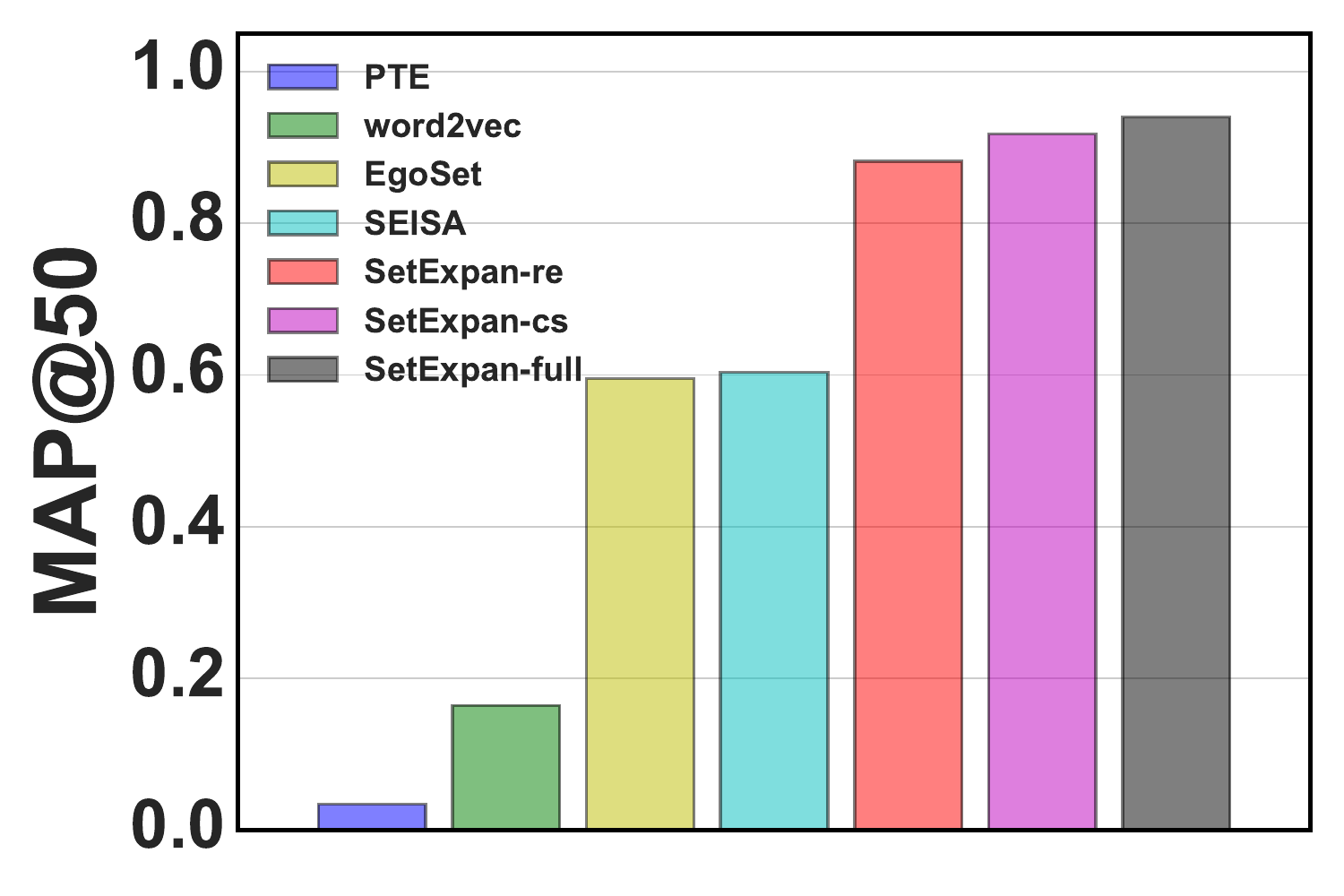}
    }
    \subfloat[\small APR Law  \label{subfig:three-types-links}]{%
      \includegraphics[width=0.33\textwidth]{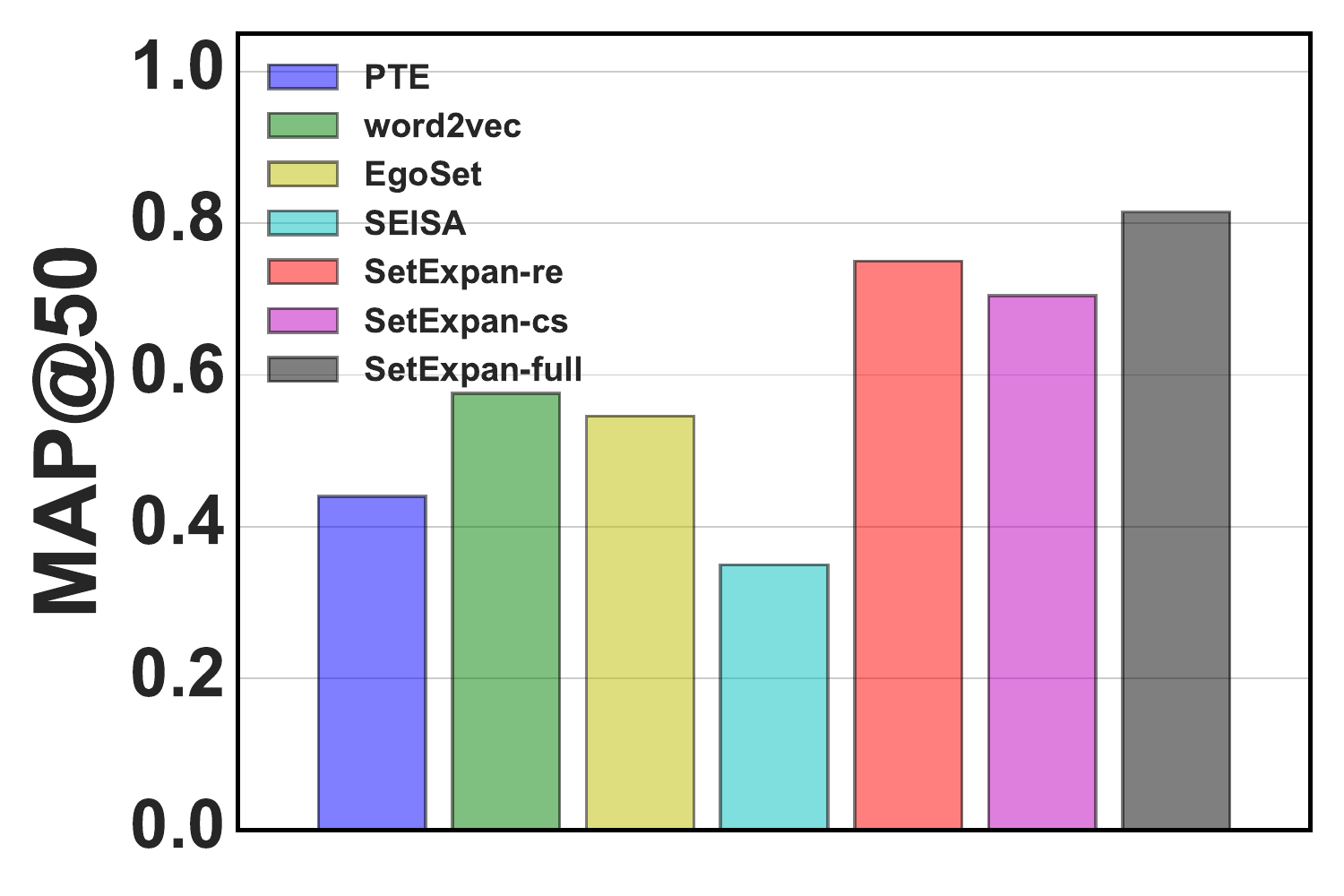}
    }
    \subfloat[\small APR Party \label{subfig:three-types-links}]{%
      \includegraphics[width=0.33\textwidth]{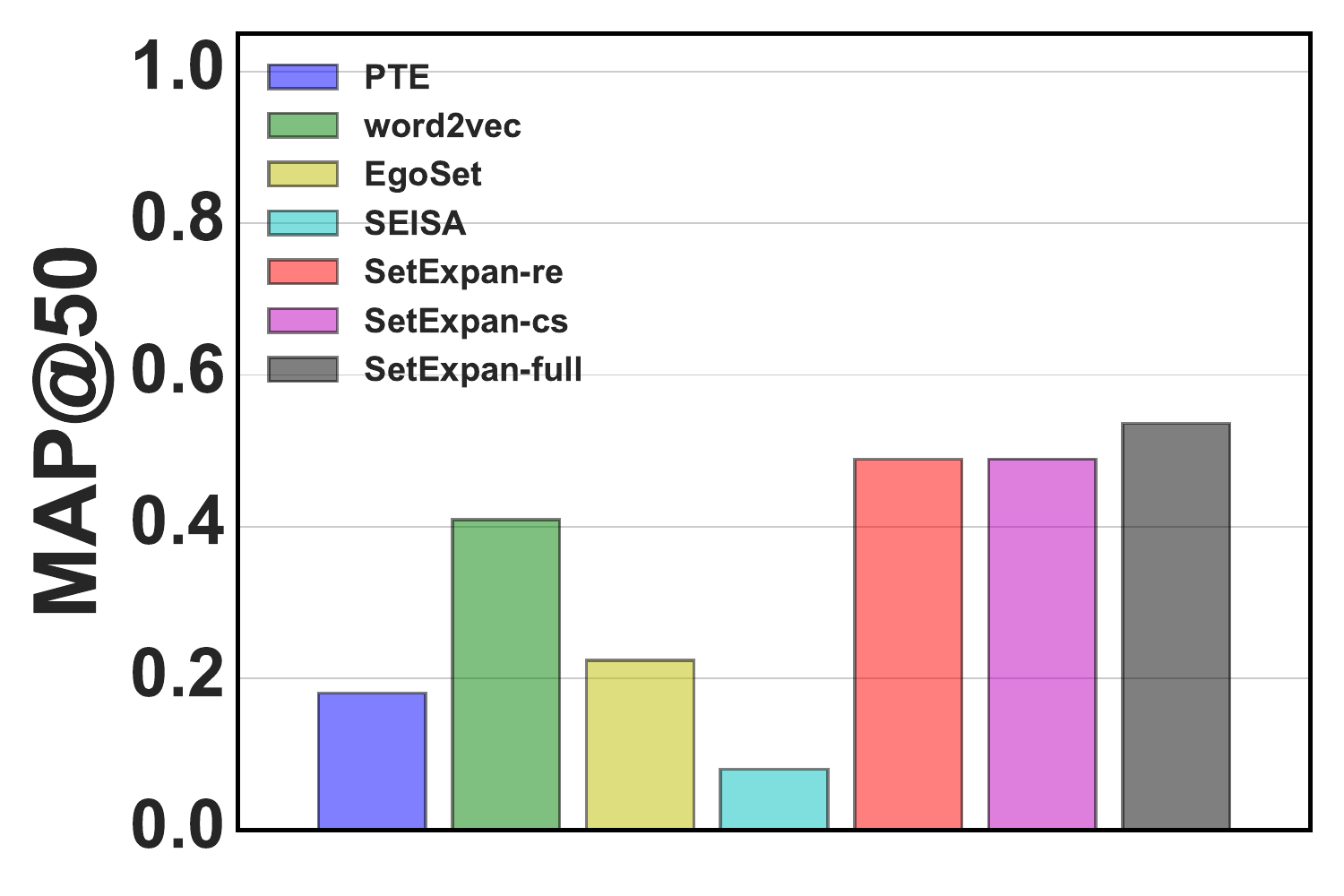}
    }
    \hfill
    \subfloat[\small Wiki Sport League \label{subfig:three-types-links}]{%
      \includegraphics[width=0.33\textwidth]{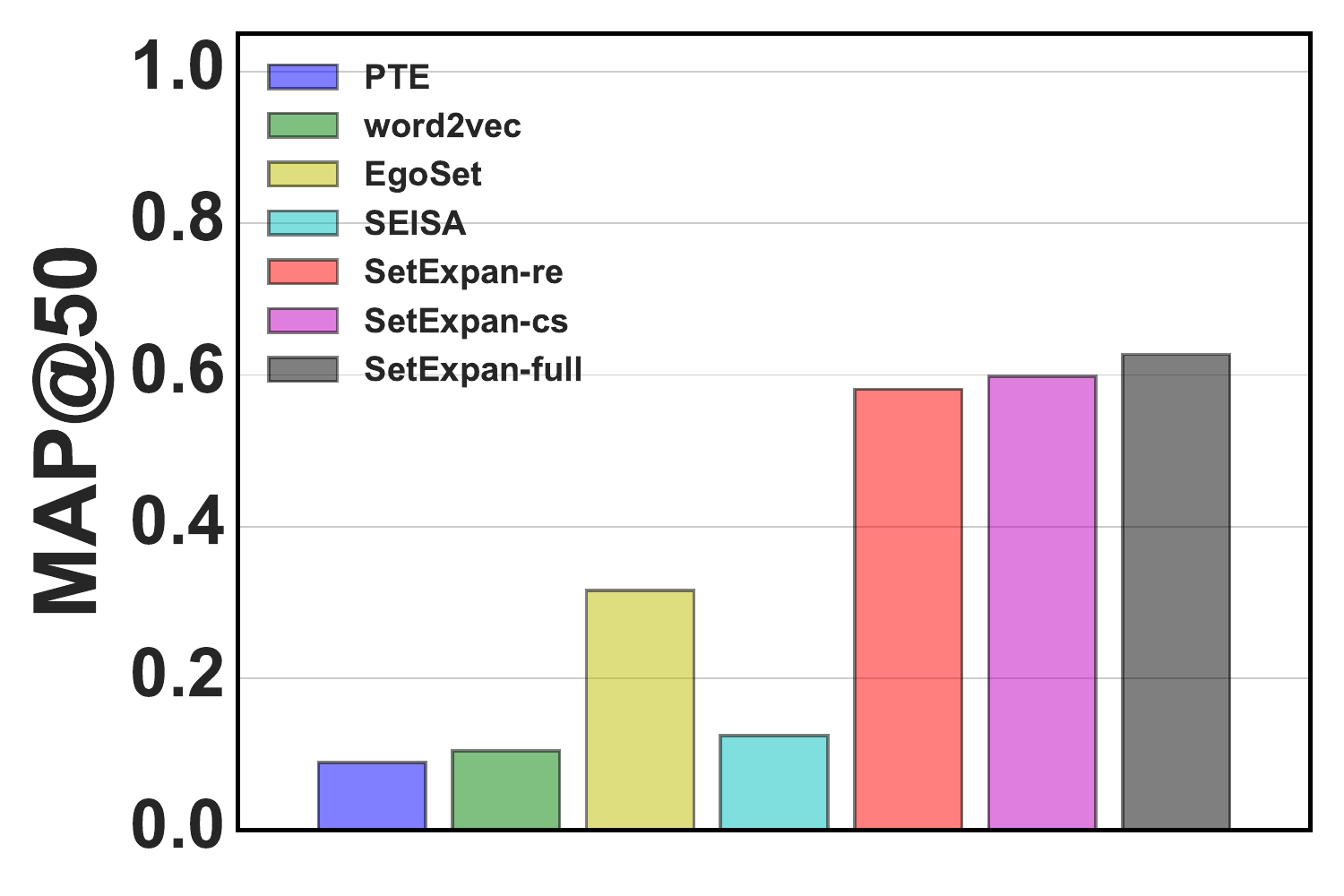}
    }
    \subfloat[\small Wiki TV Channel  \label{subfig:three-types-links}]{%
      \includegraphics[width=0.33\textwidth]{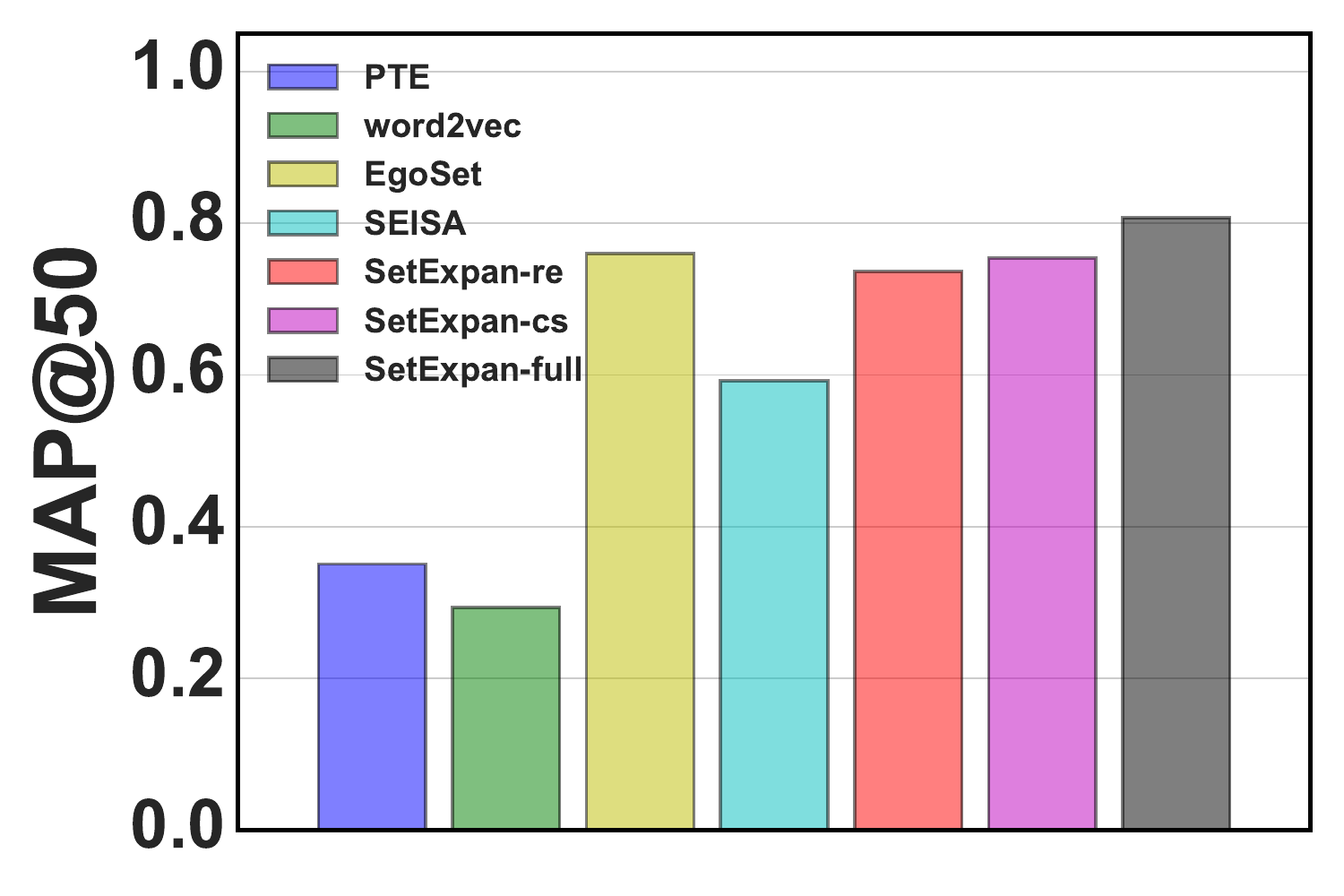}
    }
    \subfloat[\small Wiki China Province \label{subfig:three-types-links}]{%
      \includegraphics[width=0.33\textwidth]{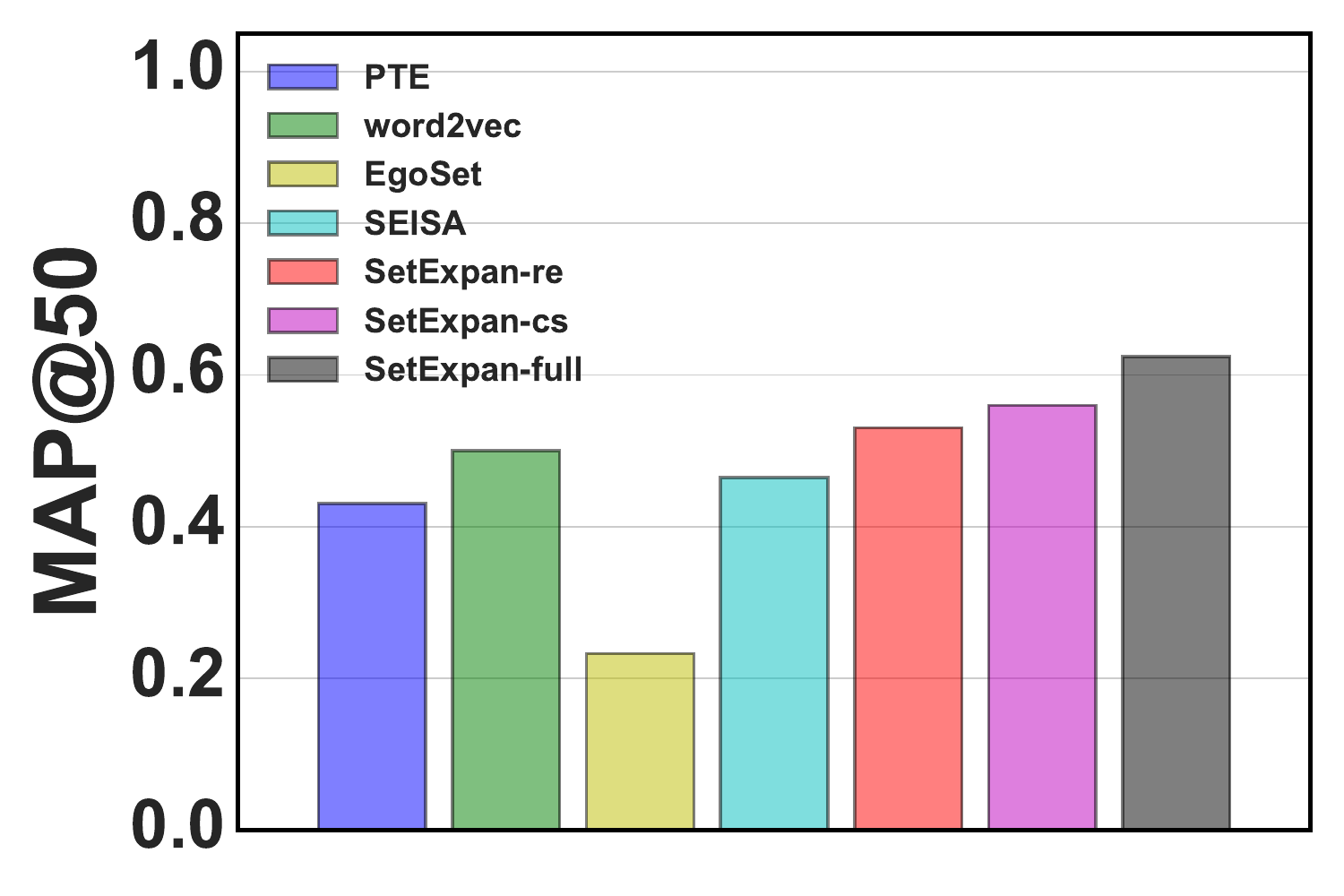}
    }
    \caption{Evaluation results for each semantic class.}
    \label{fig:conceptMAP}
  \end{figure}

%% Overall results on each query of concept class
\begin{figure}[t]
    \subfloat[\small APR Country \label{subfig:three-types-links}]{%
      \includegraphics[width=0.33\textwidth]{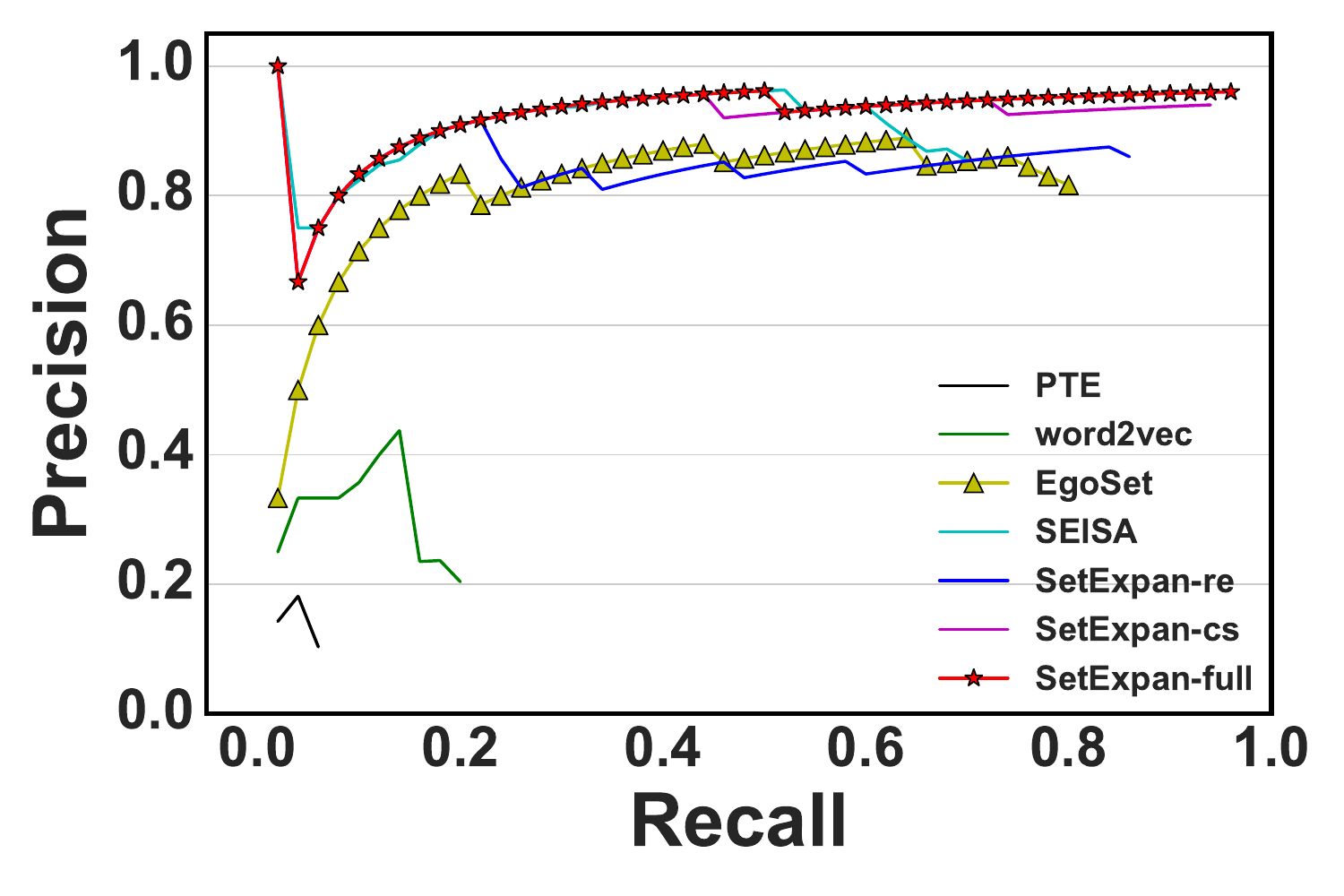}
    }
    \subfloat[\small APR Law  \label{subfig:three-types-links}]{%
      \includegraphics[width=0.33\textwidth]{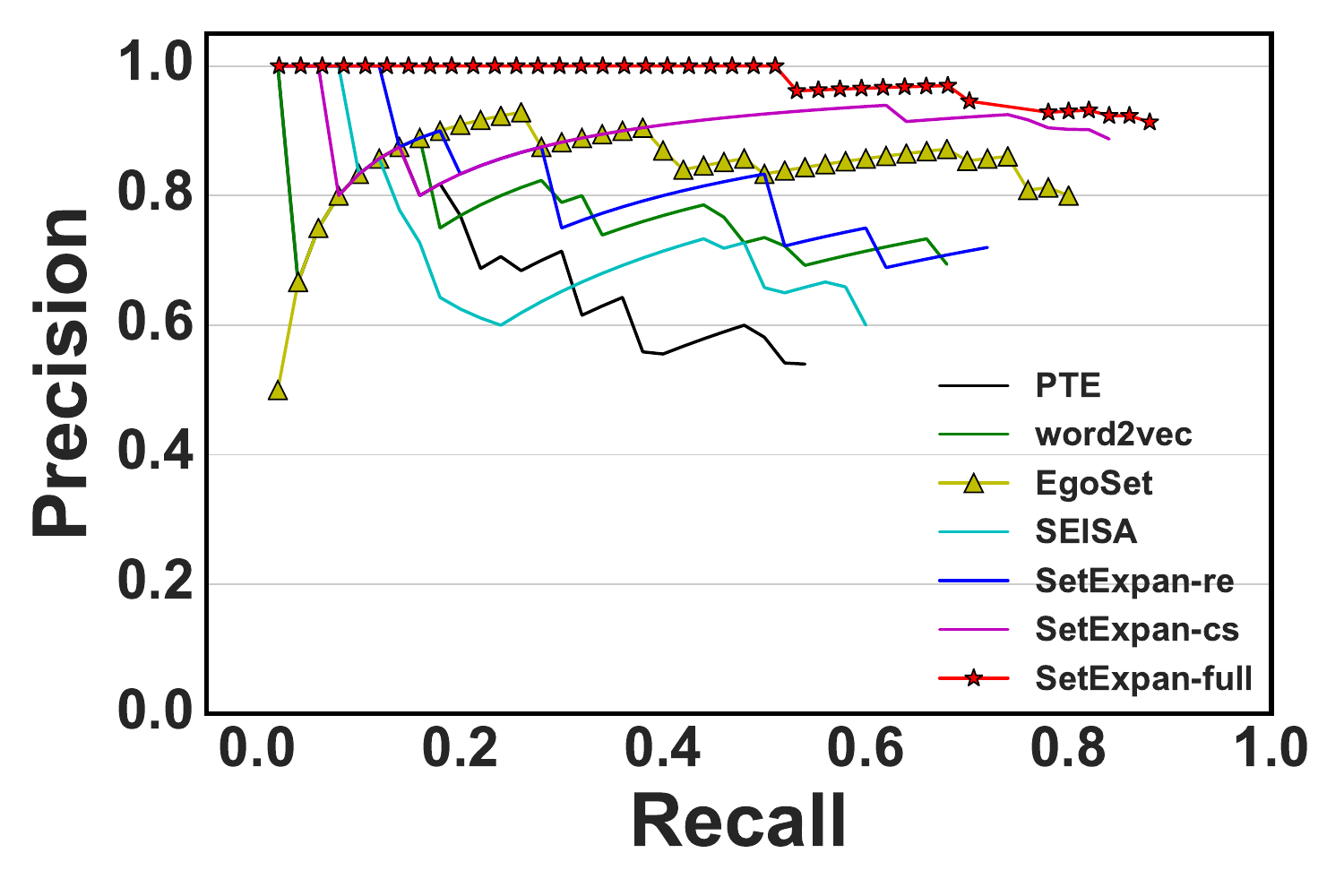}
    }
    \subfloat[\small APR Party \label{subfig:three-types-links}]{%
      \includegraphics[width=0.33\textwidth]{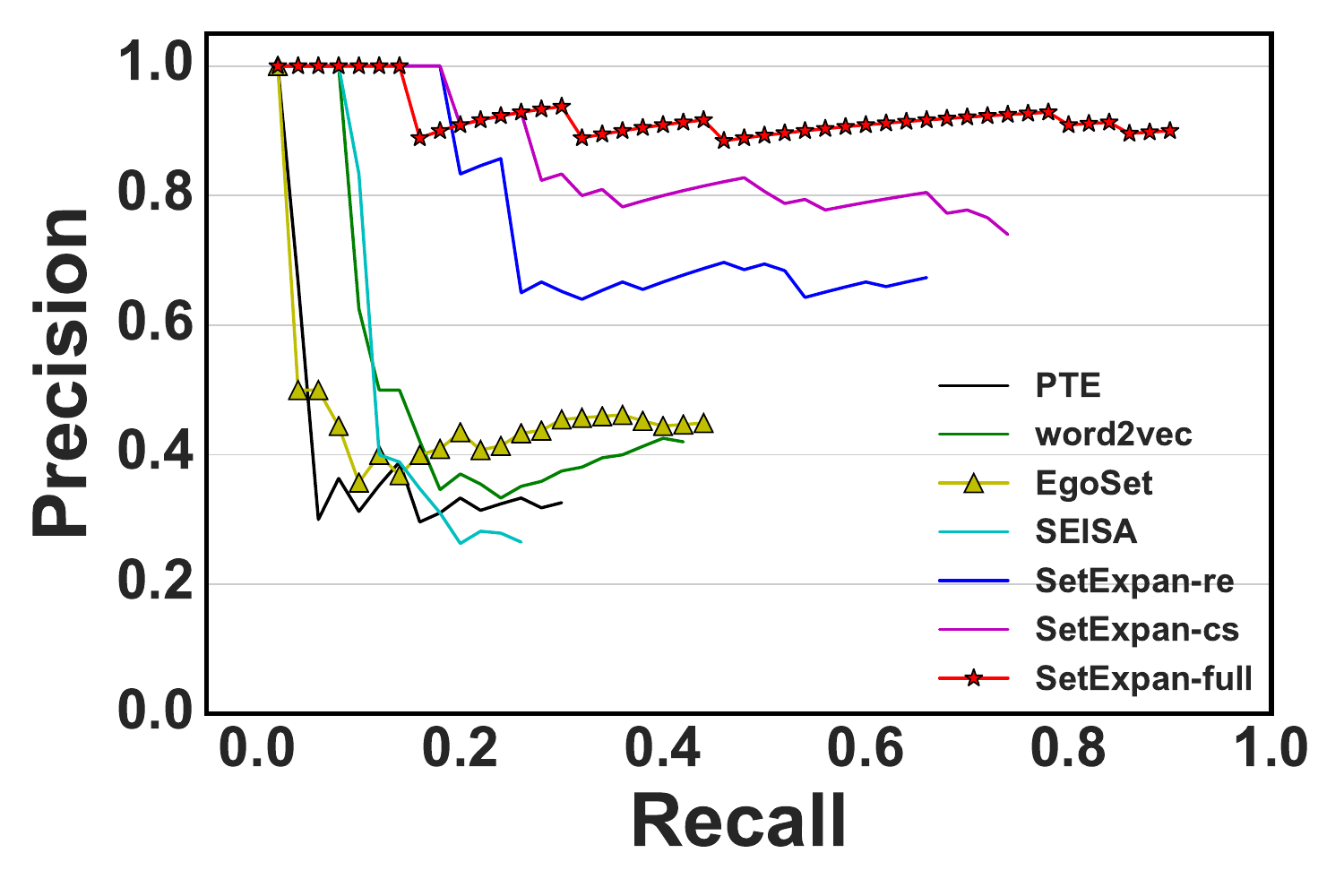}
    }
    \hfill
    \subfloat[\small Wiki Sport League \label{subfig:three-types-links}]{%
      \includegraphics[width=0.33\textwidth]{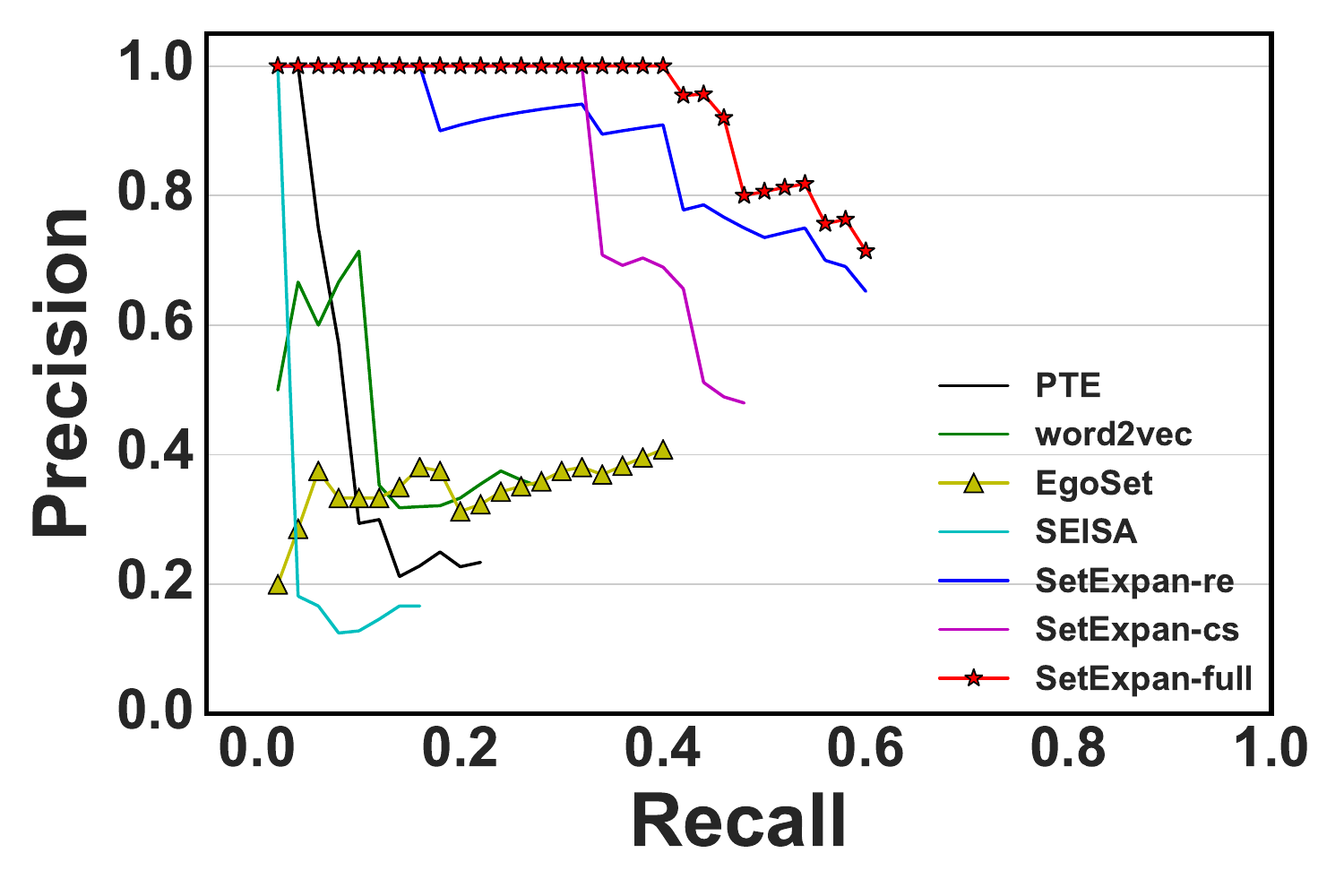}
    }
    \subfloat[\small Wiki TV Channel  \label{subfig:three-types-links}]{%
      \includegraphics[width=0.33\textwidth]{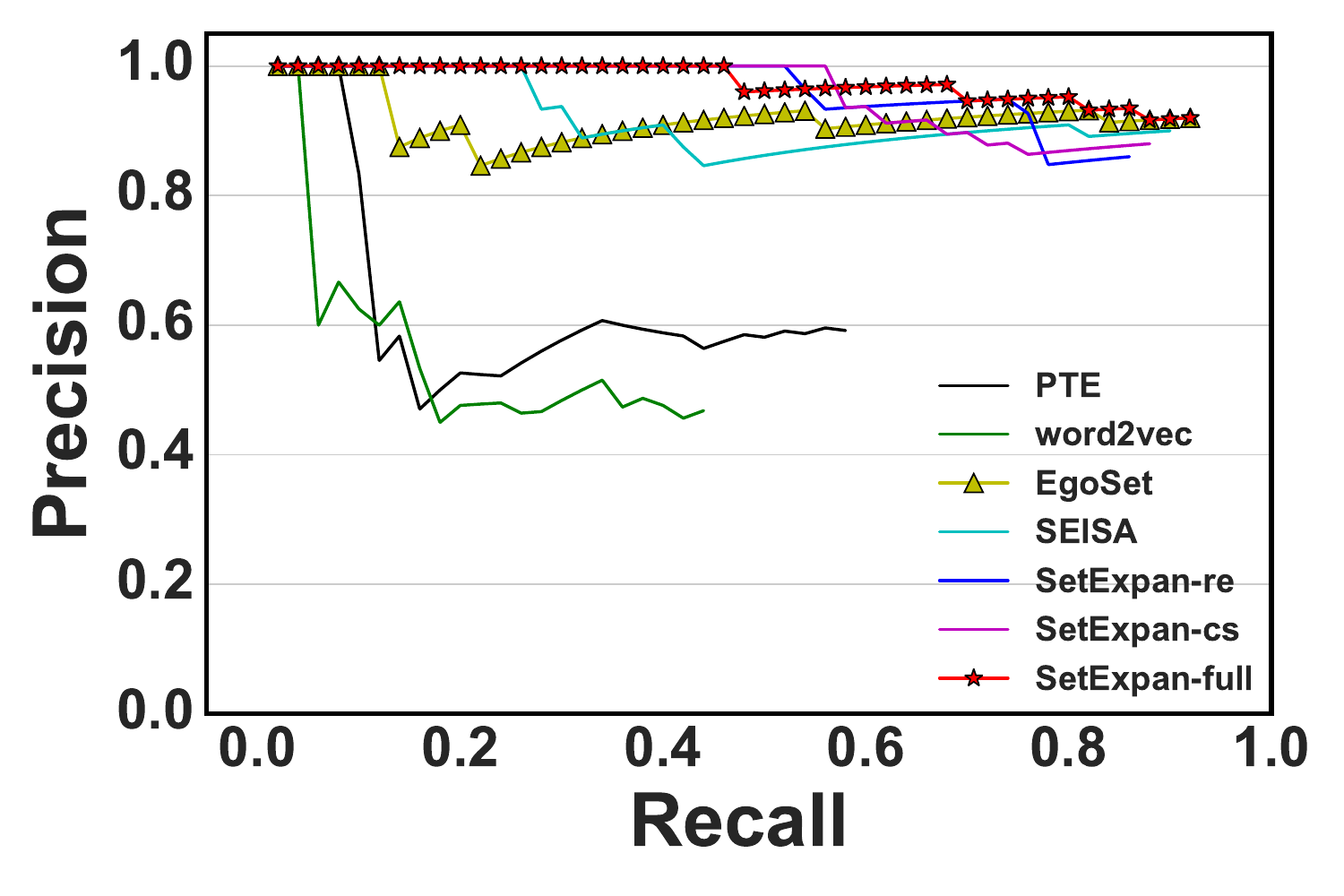}
    }
    \subfloat[\small Wiki China Province \label{subfig:three-types-links}]{%
      \includegraphics[width=0.33\textwidth]{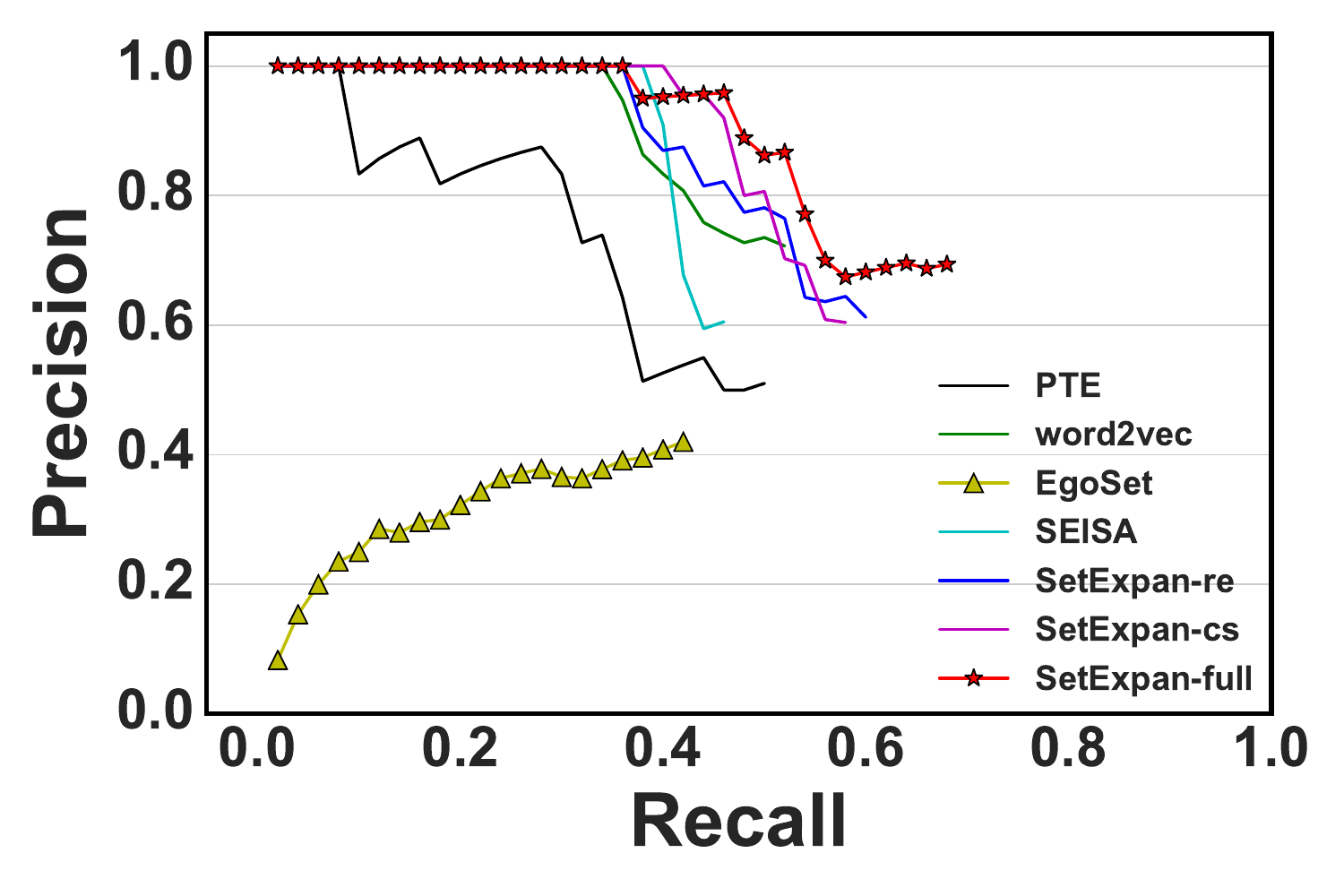}
    }
    \caption{Evaluation results for each concept class on individual query}
    \label{fig:queryPRcurve}
  \end{figure}

\subsubsection{Comparison with four baseline methods.}
Table \ref{tab:datasetMMAP} shows the MMAP scores of all methods on 3 datasets\footnote{\small Results of SEISA on PubMed-CVD are omitted due to the scalability issue.}. We can see the MMAP scores of \SetEx~outperforms all four baselines a lot.
We further look at their performances on each concept class, as shown in Figure \ref{fig:conceptMAP}.
We can see that the performance of these baseline methods varies a lot on different semantic classes, while our \SetEx~can consistently beat them.
One reason is that none of these methods applies context feature selection or rank ensemble, and a single set of unpruned features can lead to various levels of noise in the results.
Another reason is the lack of an iterative mechanism in some of those approaches.
For example, even if EgoSet includes the results from word2vec to help it boost the performance, it still achieves low MAP scores in some semantic classes.
Finding the nearest neighbors in only one iteration can be a key reason.
And although SEISA is applying the iterative technique, instead of adding a small number of new entities in each iteration, it expands a full set in each iteration based on the coherence score of each candidate entity with the previously expanded set. It pre-calculates the size of the expanded set with the assumption that the feature similarities follow a certain distribution, which does not always hold to all datasets or semantic classes.
Thus, if the size is far different from the actual size or is too big to extract a confident set at once, each iteration will introduce a lot of noise and cause semantic drift.

\textbf{Comparison with \SetExNORE~and \SetExNOCS.}
At the dataset level, the MMAP scores of \SetEx$^{\textit{full}}$~outperforms its two variation approaches.
In the semantic class level, we can see that \SetExNORE~and \SetExNOCS~sometimes have their MAP much lower than \SetEx$^{\textit{full}}$~while sometimes they almost achieve the same performance with \SetEx$^{\textit{full}}$.
This means they fail to stably extract entities with good quality.
The main reason is still that a single set of features or ensembles over unpruned features can lead to various levels of noise in the results.
Only under the circumstances that the single set of features or the unpruned features happen to be nicely selected without too much noise, which tends to happen when the query is relatively ``easy'',  these variation approaches can achieve good results.

\textbf{Effects of Context Feature Selection.}
We already see that adding the context feature selection component helps improve the performance. What's also noticeable is that the addition of context selection process becomes more obvious as the size of the corpus increases. The difference between MMAP scores of \SetExNOCS~and \SetEx$^{\textit{full}}$~is much larger in PubMed-CVD compared with APR and Wiki datasets. This is because that as the corpus size increases, we will have more noisy features and more candidate entities while the good features to define the target entity set may be limited. Thus, without context selection, noise can damage the performance much more. The evidence can also be found from the performance of EgoSet across the three datasets. It can achieve reasonably good results in APR and Wiki, however, it performs much worse in PubMed-CVD.

\textbf{Effect of Rank Ensemble.}
From the above experiments, the effect of rank ensemble has variance across the different semantic classes, however, it seems to be more stable across datasets, compared with the effect of context selection. This is because we apply the default set of parameter values in each test case above. In the parameter analysis part, we will show that the number of ensemble batches and the percentage of features to be randomly sampled can affect the contribution of rank ensemble to the set expansion performance.

\textbf{Parameter Analysis.}
There are totally 4 parameters in \SetEx~-- $Q$ (the number of selected context features), $\alpha$ (the percentage of features to be sampled), $T$ (the number of ensemble batches), and $r$ (the threshold of a candidate entity's average rank). We study the influence of each parameter by fixing all other parameters to default values, and present one graph showing the MMAP scores of \SetEx~on APR dataset versus the changes of that parameter.
\vspace{-1.5ex}
\begin{itemize}
\item $\alpha$:
From the graph, the performance increases sharply as $\alpha$ increases until it reaches about 0.6. Then, it starts to stay stable and decreases after 0.7.
\item $Q$:
In the range of 50 - 150, the performance increases sharply as $Q$ increases, which means the majority of top 150 context features can provide rich information to identify entities belonging to the target semantic class. The available information gets more and more saturated after $Q$ reaches 150 and start to introduce noises and hamper the performance after around 300.
\item $r$:
Our experiments show that the performance is not very sensitive to the threshold of a candidate entity's average rank.
\item $T$:
The performance keeps increasing as we increase the ensemble batches, due to the robustness to noise of ensembling. The performance becomes more stable after 60 batches.
\end{itemize}

%% Parameter Sensitivity Analysis
\begin{figure}[t]
    \subfloat[\small $\alpha$ \label{subfig:param-alpha}]{%
      \includegraphics[width=0.24\textwidth]{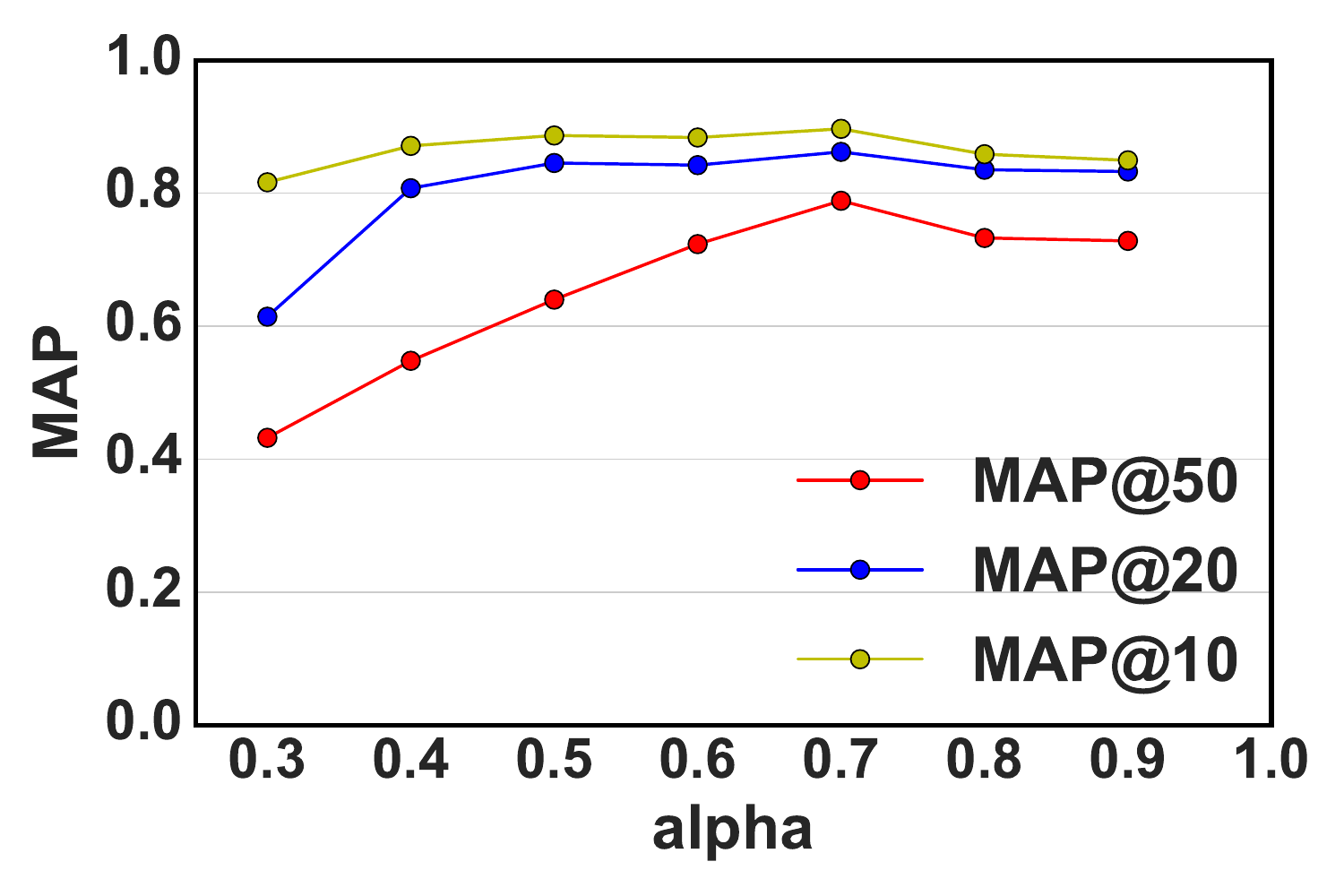}
    }
    \subfloat[\small $T$ \label{subfig:param-T}]{%
      \includegraphics[width=0.24\textwidth]{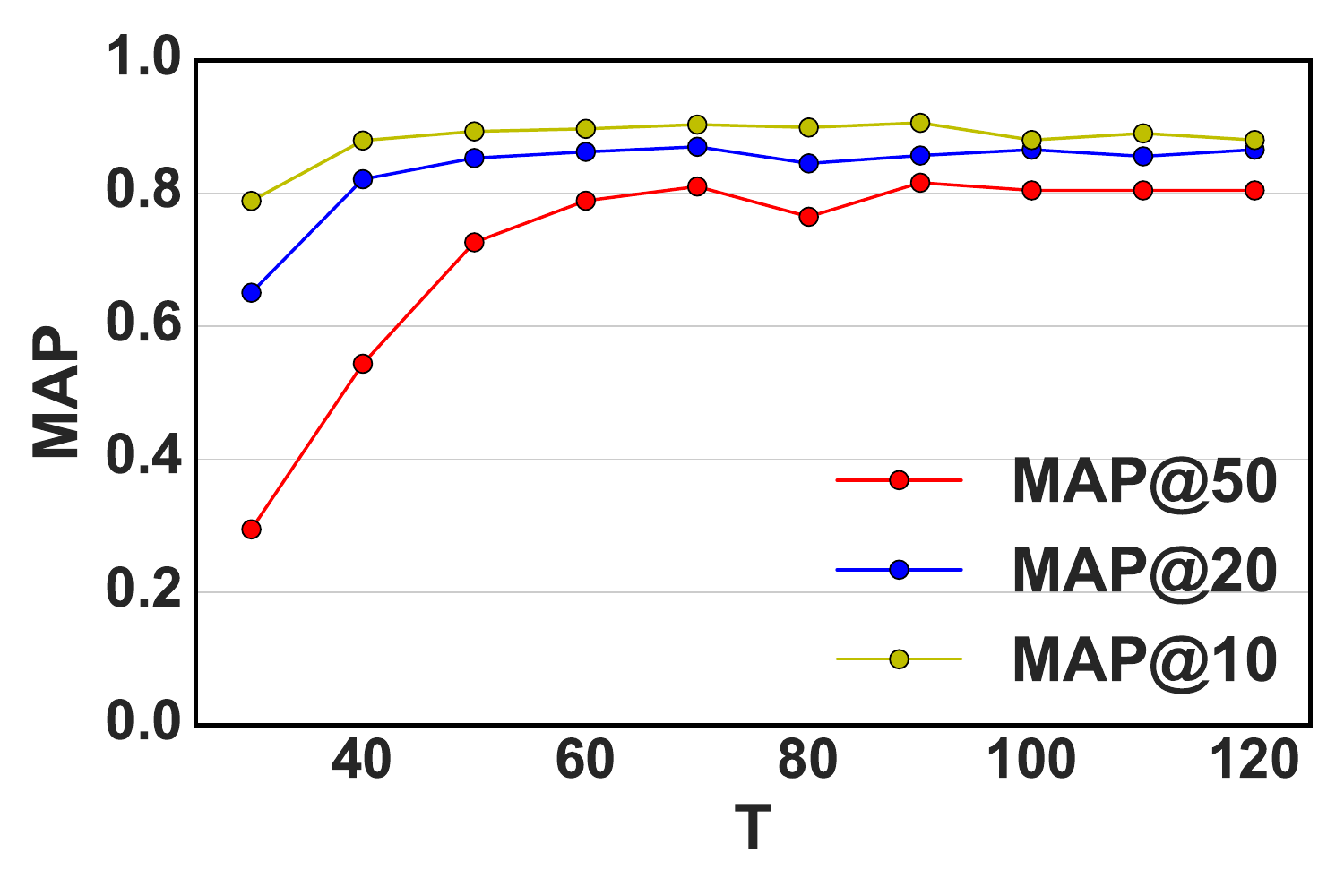}
    }
    \subfloat[\small $Q$ \label{subfig:param-Q}]{%
      \includegraphics[width=0.24\textwidth]{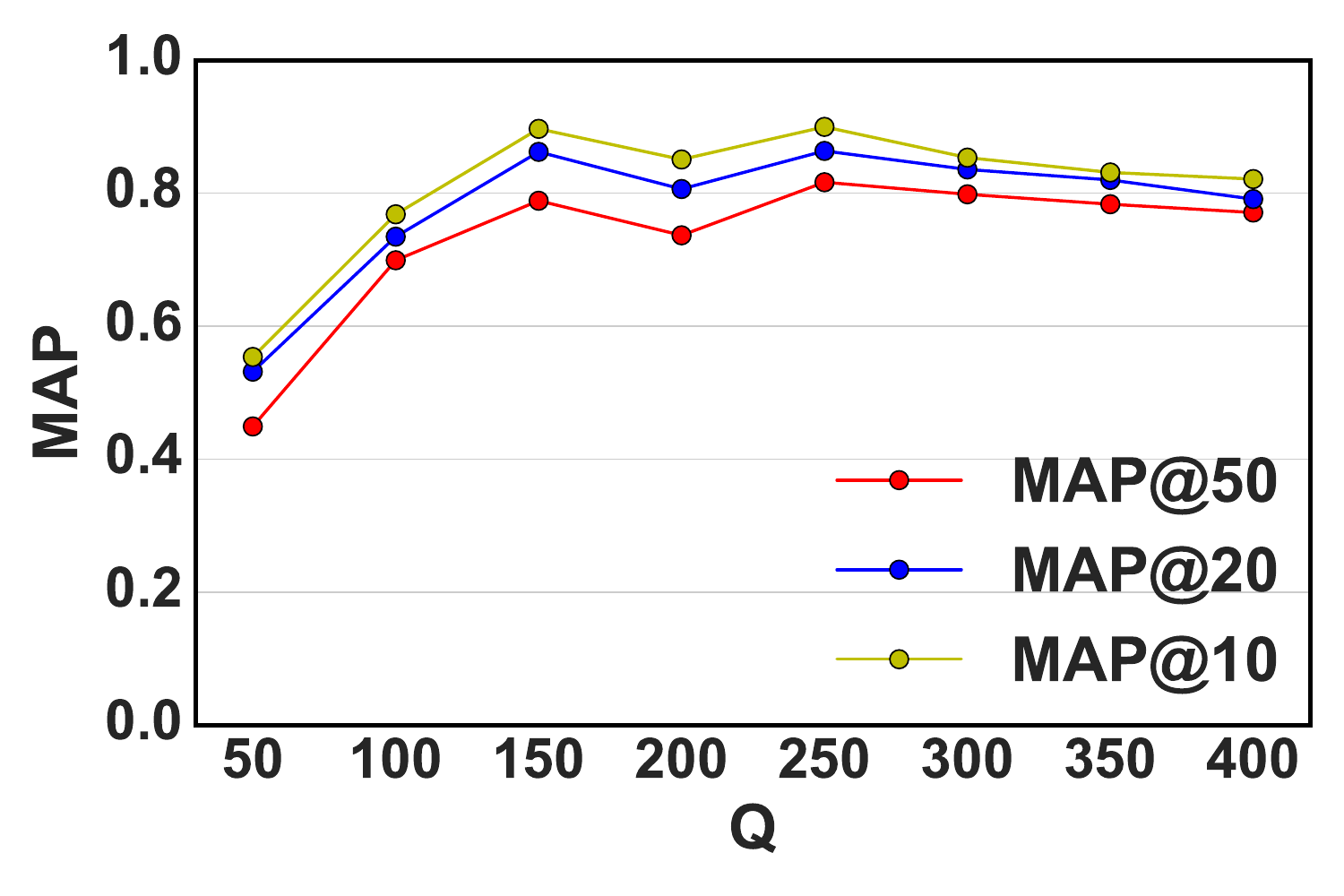}
    }
    \subfloat[\small $r$ \label{subfig:param-r}]{%
      \includegraphics[width=0.24\textwidth]{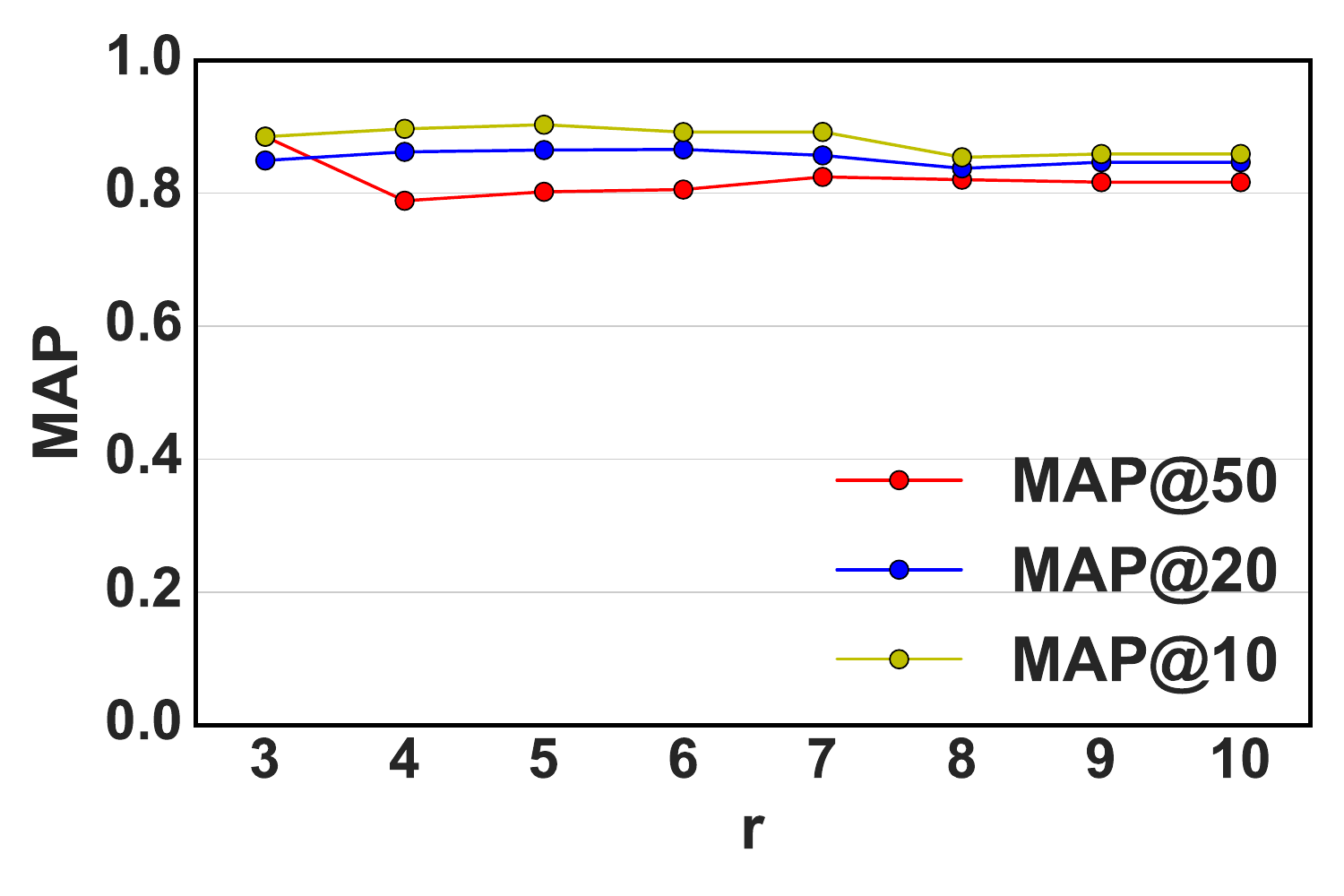}
    }
    \caption{Parameter Sensitivity on two datesets}
    \label{fig:paraSensitivityCurve}
  \end{figure}

\begin{figure}[!ht]
	\centerline{\includegraphics[width=1.00\textwidth]{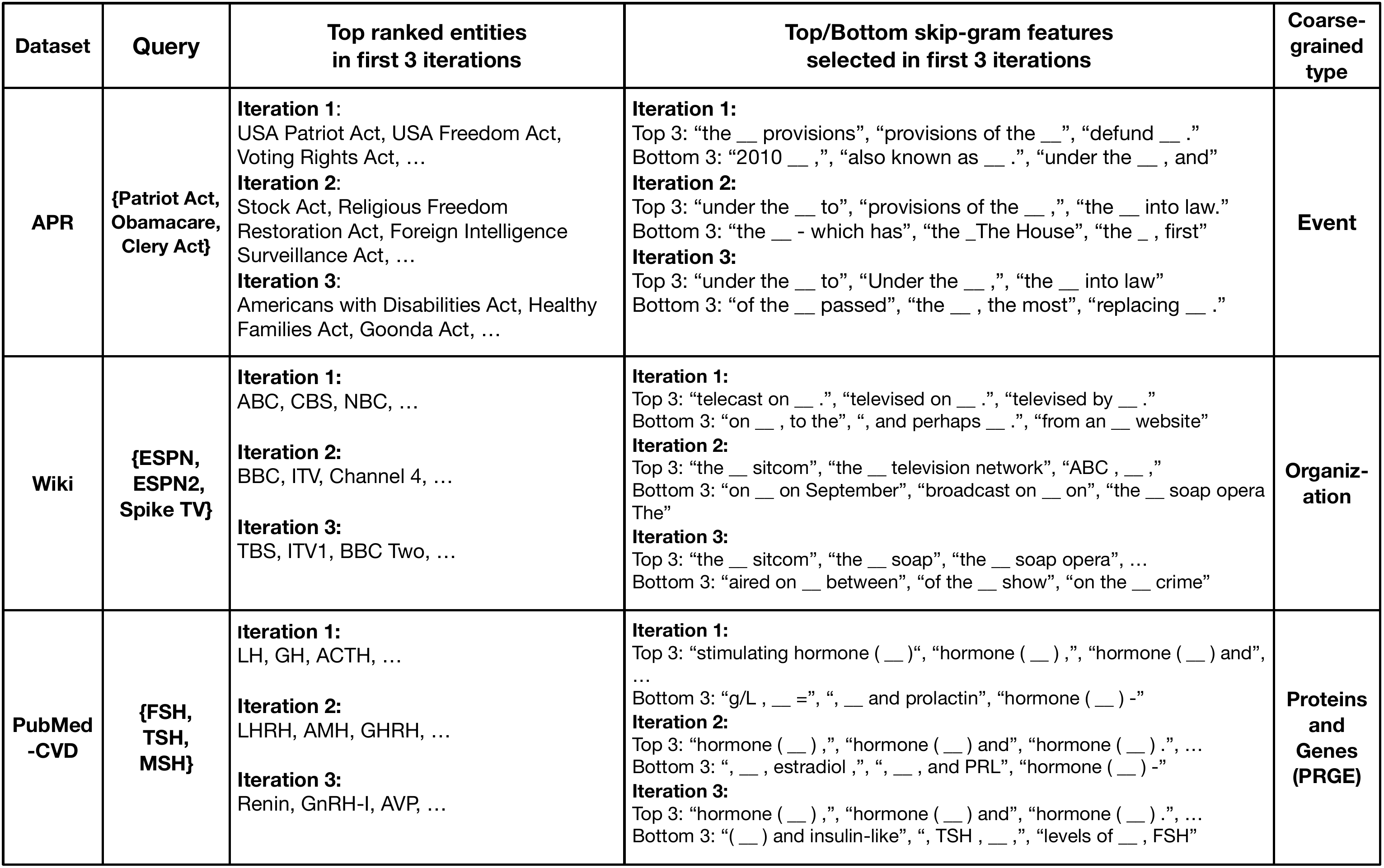}}
	\caption{Three case studies on each dataset.}
	\label{fig:casestudies}
\end{figure}

\textbf{Case Studies.}
Figure \ref{fig:casestudies} presents three case studies for \SetEx. We show one query for each dataset. In each case, we show top 3 ranked entities and top/bottom 3 skip-gram features after context feature selection for the first 3 iterations as well as the coarse-grained type. In all cases, our algorithm successfully extracts correct entities in each iteration, and the top-ranked skip-grams are representative in defining the target semantic class. On the other hand, we notice that most of the bottom 3 skip-grams selected are very general or not representative at all. These context features could potentially introduce noisy entities and thus the rank ensemble can play a rival role in improving the results.

%!TEX root = main.tex
% UTF-8 encoding
\vspace{-1.0ex}
\section{Conclusion and Future Work} 
\vspace{-0.5ex}
In this paper, we study the problem of corpus-based set expansion. First, we propose an iterative set expansion framework with a context feature selection method, to deal with the problem of entity intrusion and semantic drift. Second, we develop a novel unsupervised ranking-based ensemble algorithm for entity selection, to further reduce context noise in free-text corpora. Experimental results on three publicly available datasets corroborate the effectiveness and robustness of our proposed \SetEx. 

The proposed framework is general and can incorporate other context features besides skip-grams, such as Part-Of-Speech tags or syntactic head tokens. Besides, it would be interesting to study more rank ensemble methods for aggregating multiple pre-ranked lists. In addition, our current framework treats each feature independently, it would be interesting to study how the interaction of context features can influence the expansion result. We leave it for future work.
\smallskip
\subsubsection{Acknowledgments.} Research was sponsored in part by the U.S. Army Research Lab. under Cooperative Agreement No. W911NF-09-2-0053 (NSCTA), National Science Foundation IIS-1320617, IIS 16-18481, and NSF IIS 17-04532, and grant 1U54GM114838 awarded by NIGMS through funds provided by the trans-NIH Big Data to Knowledge (BD2K) initiative (www.bd2k.nih.gov).

\vspace{-1.0ex}

%NOTE: Double-check the reference format before the submission
\bibliographystyle{abbrv}
\scriptsize
\bibliography{pkdd2017}

\end{document}